\documentclass[Afour,sageh,times]{sagej}

\def\volumeyear{2025}
\usepackage{moreverb,url}  
\usepackage[colorlinks,bookmarksopen,bookmarksnumbered,citecolor=red,urlcolor=red]{hyperref}  

\usepackage{graphicx}
\usepackage{amsfonts}
\usepackage{mathtools}  
\usepackage{bm}  
\usepackage{caption}  
\usepackage{xspace}  
\usepackage{booktabs}  
\usepackage{makecell} 
\usepackage{multirow} 
\usepackage{bigdelim}
\usepackage[dvipsnames,table]{xcolor}  

\DeclareMathOperator*{\argmax}{\arg\!\max}
\newcommand{\norm}[1]{\left\lVert#1\right\rVert}   
\newcommand{\vect}[1]{\mathbf{#1}}  
\newcommand{\mat}[1]{\vect{#1}}  
\newcommand{\vecg}[1]{\bm{#1}}  
\newcommand{\R}{\mathbb{R}}  
\newcommand{\bmat}[1]{\begin{bmatrix} #1 \end{bmatrix}}  
\newcommand{\smat}[1]{\left[\begin{smallmatrix} #1 \end{smallmatrix}\right]} 
\newcommand{\inv}{^{-1}}  
\newcommand{\transp}{^{\top}}  
\newcommand{\ts}[1]{{}^{#1}}  
\newcommand{\us}[1]{{}_{#1}}  
\newcommand{\frameV}[1]{\mathcal{F}_{#1}}  

\newcommand{\interv}[2]{\left[#1,\;#2\right]}  
\newcommand{\intervexcl}[2]{\left(#1,\;#2\right)}  

\newcommand{\ie}{i.e.\xspace}
\newcommand{\eg}{e.g.\xspace}

\newcommand{\degree}{$^{\circ}$\xspace}
\newcommand{\toptext}[1]{${}^{\text{#1}}$\xspace}
\newcommand{\ith}{\textit{-th}\xspace}

\newcommand{\fig}[1]{Figure~\ref{#1}\xspace}
\newcommand{\sect}[1]{Section~\ref{#1}\xspace}
\newcommand{\appen}[1]{Appendix~\ref{#1}\xspace}
\newcommand{\tab}[1]{Table~\ref{#1}\xspace}
\newcommand{\eqn}[1]{Equation~\eqref{#1}\xspace}

\newcommand{\tturl}[1]{\tt\url{#1}}
\newcommand{\footurl}[1]{\footnote{\noindent\tturl{#1}}}

\newcommand{\num}[1]{#1}
\newcommand{\unit}[1]{\ensuremath{\mathrm{#1}}}
\newcommand{\si}[1]{\ensuremath{\mathrm{#1}}}
\newcommand{\SI}[2]{\ensuremath{#1\,\mathrm{#2}}}

\begin{document}

\runninghead{Jacquet et al.}

\title{Neural NMPC through Signed Distance Field Encoding for Collision Avoidance}

\author{Martin Jacquet, Marvin Harms and Kostas Alexis}

\affiliation{All authors are affiliated with the Autonomous Robots Lab, Norwegian University of Science and Technology (NTNU), Trondheim, Norway}

\corrauth{Martin Jacquet,
Autonomous Robots Lab,
Department of Engineering Cybernetics,
Norwegian University of Science and Technology,
Høgskoleringen 1, 7034 Trondheim, Norway.}

\email{martin.jacquet@ntnu.no}

\begin{abstract}
This paper introduces a neural Nonlinear Model Predictive Control (NMPC) framework for mapless, collision-free navigation in unknown environments with Aerial Robots, using onboard range sensing.
We leverage deep neural networks to encode a single range image, capturing all the available information about the environment, into a Signed Distance Function (SDF).
The proposed neural architecture consists of two cascaded networks: a convolutional encoder that compresses the input image into a low-dimensional latent vector,
and a Multi-Layer Perceptron that approximates the corresponding spatial SDF.
This latter network parametrizes an explicit position constraint used for collision avoidance,
which is embedded in a velocity-tracking NMPC that outputs thrust and attitude commands to the robot.
First, a theoretical analysis of the contributed NMPC is conducted, verifying recursive feasibility and stability properties under fixed observations.
Subsequently, we evaluate the open-loop performance of the learning-based components as well as the closed-loop performance of the controller in simulations and experiments.
The simulation study includes an ablation study, comparisons with two state-of-the-art local navigation methods, and an assessment of the resilience to drifting odometry.
The real-world experiments are conducted in forest environments, demonstrating that the neural NMPC effectively performs collision avoidance in cluttered settings against an adversarial reference velocity input and drifting position estimates.

\end{abstract}


\maketitle

\section{Introduction}\label{sec:intro}

\begin{figure*}[t]
    \includegraphics[width=\linewidth]{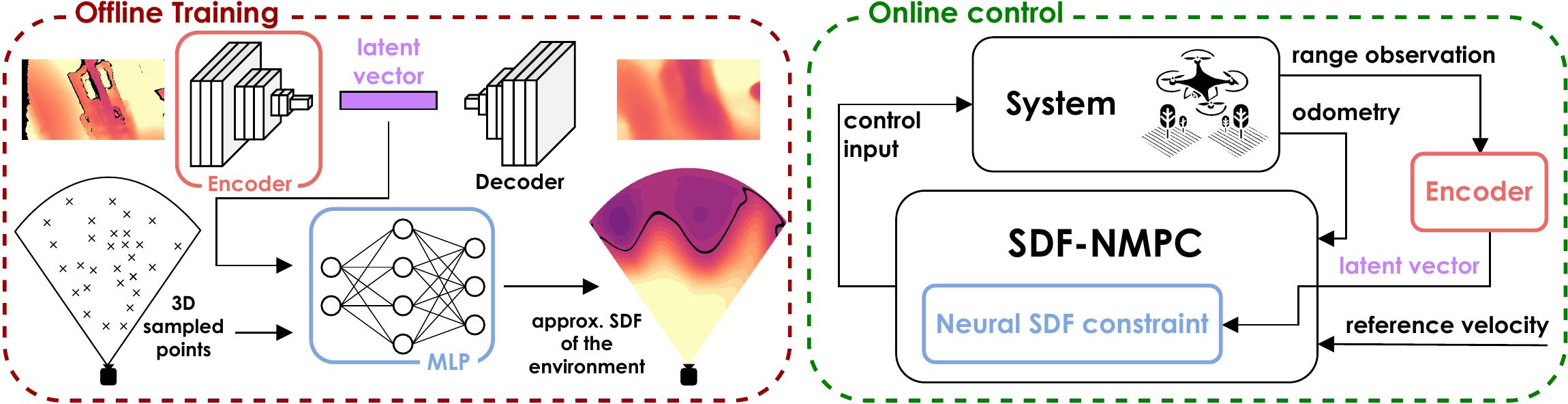}
    \caption{An overview of the proposed SDF-NMPC method.
        The left side depicts the neural architecture used to approximate the mapping between the input depth images and the corresponding SDF, through sampling-based training in the 3D sensor frustum.
        The Convolutional encoder-decoder and MLP networks are trained sequentially.
        The right side presents the proposed control scheme, highlighting the contributed neural-constrained NMPC for velocity tracking, and the color-coded learning-based components.}
    \label{fig:intro:overview}
\end{figure*}

Recent advances in autonomous navigation and aerial robotics have enabled the large-scale deployment of robots in several challenging applications, such as subterranean or forest exploration and ship inspection~\cite{Fang17, Tian20, Tranzatto22}.
However, navigation in unknown and unstructured environments while relying only on onboard sensing remains a strenuous challenge.
This is particularly true in perceptually-degraded environments and conditions, where mapping systems are failure-prone ~\cite{Ebadi23}.

Conventional state-of-the-art navigation stacks often involve the construction of a reliable volumetric map of the scene, followed by a planning step~\cite{Sucan12, Tranzatto22}.
Although this approach has proven effective in many cases, errors at the mapping level (such as a sudden localization drift) propagate downstream, adversely affecting planning and control.
Furthermore, the sequential update of dense maps before motion planning induces latency, which limits the responsiveness to unforeseen events.
Enhanced resilience w.r.t. collisions can be achieved by integrating map-based methodologies with local reactive strategies that rely on instantaneous exteroceptive observations.

Accordingly, the robotics community has investigated a set of approaches for the so-called mapless navigation problem,
\ie, the problem of achieving collision-free navigation without relying on explicit mapping.
Beyond early heuristic approaches to reactive collision avoidance~\cite{Siegwart11}, more advanced strategies have emerged in recent years.
While initial efforts focused on model-based methods~\cite{Florence18, Gao19, Florence20, Yadav20, Zhou21a},
deep learning-based techniques have taken the field by storm, primarily due to their scalability to high-dimensional exteroceptive inputs such as depth images.
Relevant learning strategies include Imitation Learning (IL)~\cite{Loquercio21, Lu23},
and Reinforcement Learning (RL)~\cite{Hoeller21, Ugurlu22, Kulkarni24}.
In addition to these end-to-end methods, recent research has investigated the use of deep learning for implicit environment representations,
combined with principled control strategies~\cite{Harms24, Jacquet24}.

This work falls into this latter category, and its main contribution is threefold.
First, we introduce a neural architecture for encoding a single range observation as an Signed Distance Function (SDF)~\cite{Park19, Sitzmann20}.
Unlike prior work on (neural or non-neural) scene-scale implicit SDF encoding -- which primarily targets mapping and planning applications~\cite{Ortiz22, Wu23} -- we explicitly avoid aggregating multiple observations.
Second, we integrate this representation into a neural Nonlinear Model Predictive Control (NMPC) for achieving collision-free, mapless navigation in unknown environments.
We verify that the NMPC satisfies recursive feasibility (under fixed observations) and stability conditions.
An overview block diagram of the contributed method, referred to as SDF-NMPC, is shown in~\fig{fig:intro:overview}.
Thirdly, we conduct a detailed, component-wise evaluation, including a
quantitative assessment of the neural environment encoding, ablation studies,
comparisons with existing methods, and real-world experiments involving drifting odometry and adversarial reference inputs.

The proposed NMPC relies solely on instantaneous range measurements from an onboard sensor and on odometry, which may suffer from position drift.
The neural SDF encoding of the current range observation parametrizes a position constraint for the controller.
To address the limited sensor Field of View (FoV), the predicted motion is also constrained to remain within the visible frustum.
The controller generates thrust and attitude commands that satisfy both collision and FoV constraints, and is integrated in a velocity-tracking framework for real-time control of an Aerial Robot (AR).
The implementation of the contributed NMPC and the associated neural networks is released open-source\footurl{https://github.com/ntnu-arl/sdf-nmpc}.

Accordingly, this work departs from several limitations of our previous contribution in~\cite{Jacquet24}.
First, the proposed control scheme generalizes to the nonlinear dynamics of the multirotor, moving beyond the previously considered unicycle-like motions.
Second, we make use of a richer 3D representation. Indeed, the SDF is well-suited for collision avoidance, being a continuous, almost everywhere differentiable 3D field that encodes both the distance and direction to the nearest obstacle.
Defined purely at the position level, it can be constructed directly from sensor measurements, whereas defining safe sets for derivative states (e.g., velocity or acceleration) is often challenging for nonlinear, higher-order systems~\cite{Harms24}.
Moreover, the neural encoding provides a closed-form differentiable expression, enabling straightforward integration with gradient-based NMPC.
The SDF formalism also facilitates the definition of a forward-invariant safe set for the receding-horizon problem, allowing us to establish recursive feasibility of the control law without introducing excessively restrictive terminal constraints.
Lastly, the evaluation is substantially more comprehensive, including the explicit evaluation of the resilience of the $0$-memory approach against drifting odometry.

The remainder of this article is structured as follows.
\sect{sec:relatedwork} provides an overview of the related literature,
followed by the problem statement in \sect{sec:problem}.
The proposed method is presented in \sect{sec:nn} and \sect{sec:mpc}, respectively covering the neural environment encoding and the collision-avoidance NMPC framework.
Finally, \sect{sec:valid} presents a comprehensive evaluation of the method, before discussions and final remarks in \sect{sec:ccl}.

\section{Related Work}\label{sec:relatedwork}

This section presents related work, and especially
a)~recent advances in neural NMPC,
b)~methodologies for mapless navigation, and
c)~neural distance fields and their applications in robotics.

\subsection{Neural MPC}

Nonlinear Model Predictive Control (MPC) has has become a widely adopted control strategy for constrained systems, showing strong performance for ARs~\cite{Sun22}.

With advances in onboard computing and machine learning,
learning-based MPC emerged as a successful alternative to robust or stochastic MPC, offering online adaptability to model mismatches and external disturbances~\cite{Hewing20}.
Gaussian Processes, for instance, have been used to model residual dynamics in~\cite{Kabzan19,Torrente21}.
Neural Networks (NNs), capable of capturing complex data patterns, have also been used to model complex dynamics,
such as aerodynamic effects for quadrotor flight~\cite{Bauersfeld21}.
Following a similar approach, neural NMPC has been proposed,
embedding small-scale NNs to learn unmodeled dynamics~\cite{Williams17,Chee22,Gao24}.
Going further,~\cite{Syntakas24} introduces an ensemble approach based on the Monte-Carlo dropout technique, addressing the epistemic uncertainty in the neural models.
RL has also been used to optimize the NMPC model, cost, and constraints to enhance closed-loop performance\cite{Gros19}, further extended by including a neural model for unknown dynamics~\cite{Adhau24}.

An open-source toolbox for integrating large NNs into NMPC is proposed in~\cite{Salzmann23},
enabling tasks like navigation in turbulent flow~\cite{Salzmann24}.
This paved the way for exploiting complex neural representations in NMPC.
In~\cite{Alhaddad24}, a transformer-based NN have been used to represent obstacle fields as repulsive cost,  while~\cite{Jacquet24} embeds depth camera observations into a differentiable occupancy map, explicit used as a position constraint.
This work builds upon this architecture.

Other recent architectures reframe NMPC as a parameterized controller learned via policy search for agile flight~\cite{Song22}, or integrate differentiable MPC into RL agents for tighter coupling between short- and long-term control~\cite{Romero24}. RL-based warm-starting strategies have been proposed to initialize the NMPC with actor-critic rollouts~\cite{Reiter24}, or approximate terminal costs using neural networks~\cite{Alsmeier24}. Additionally,~\cite{Celestini24} leverages a Transformer to generate trajectory candidates for warm-starting the solver.

A set of novel architectures for neural MPC has been explored.
In~\cite{Song22}, the NMPC is formulated as a parametrized controller learned via policy search for agile flight.
Another avenue relies on NN to approximate the infinite horizon cost.
In ~\cite{Romero24}, a differentiable MPC implementation~\cite{East20} is integrated in the RL agent, such that the MPC drives the short-term actions while the critic network manages the long-term ones.
RL-based warm-starting strategies have been proposed in~\cite{Reiter24} to initialize the NMPC via policy roll-out.
In~\cite{Alsmeier24}, the long-horizon cost of the MPC is approximated by the NN, and used as the terminal cost of a short-horizon problem, for computational efficiency.
Additionally, in~\cite{Celestini24}, a Transformer is used to generate candidate trajectories for warm-starting MPC solver.

\subsection{Mapless Navigation}

Mapless navigation has gained attention in recent years.
A first class of approaches constructs local volumetric representations from recent observations using structures like k-d trees~\cite{Florence18, Gao19}, enabling fast collision checking.
Ray-casting~\cite{Yadav20},
or depth map back-projection~\cite{Matthies14} can also be used for fast collision checks.
Leveraging such efficient checks,  motion primitives have been used for planning~\cite{Lopez17, Bucki20}.
In~\cite{Zhou21a}, a spline-based planner using differentiable repulsive forces from visible obstacles is introduced.
Contrary to these works, which rely on a single depth measurement to perform checks,~\cite{Florence18} introduces a data representation for querying points in a sliding window of consecutive observations, explicitly accounting for the transform uncertainty induced by the uncertain state estimation.
In~\cite{Zhang25}, a NMPC is proposed leveraging a k-d tree to query the $N$ closest obstacles for each shooting node along a pre-sampled path, that are in turn used as $1$D position constraints for this shooting node.
The approach struggles in cluttered settings where the path must significantly diverge from the straight-line reference.

Deep learning offers an alternative to collision checking, and exhibits good results for mapless navigation.
\cite{Nguyen22, Nguyen24} proposes a neural architecture to correlate depth images to near-future collisions for a given action.
This method was extended with a Variational Auto-Encoder (VAE) to allow more control over the latent representation, enabling semantic augmentation for task-specific navigation~\cite{Kulkarni23a}.
A similar approach is proposed in~\cite{Lu23} where a Q-value-inspired model predicts action values using expert policy rollouts.
In~\cite{Liang24}, a generative NN is used to generate candidate trajectories that satisfy traversability and coverage learned from A*-generated data.

Imitation Learning (IL) has also been employed to learn how to generate sensor-based trajectories by imitating a map-informed export policy, such as an NMPC~\cite{Tolani21}, a sampling-based expert~\cite{Loquercio21}, or a privileged RL agent~\cite{Song23}.
However, IL often lacks generalization of the learned policy.
Thus, numerous works tackled this problem with RL~\cite{Kahn21a,Kahn21b}.
For instance,~\cite{Ugurlu22} trains a policy from depth data but neglects AR dynamics during training, while~\cite{Kulkarni24} incorporates dynamics and achieves real-world deployment.
Some RL-based approaches also learn implicit representations of time-consistent environments and robot state~\cite{Hoeller21}.

A hybrid strategy leverages NNs to encode a compressed, interpretable environment representation.
Several works use neural networks to learn Control Barrier Functions (CBFs) for a given system, which describes safety w.r.t. to the obstacles as described in range measurements.
A relevant instance is provided by~\cite{Dawson22}, where an observation-based CBF is learned for a first-order system.
The method relies on approximating the LiDAR observation dynamics, which is difficult for higher-order systems and error-prone in complex environments, because the true observation dynamics are, in general, discontinuous.
More accurate prediction of the next observation can be achieved with a NeRF~\cite{Tong23} and used in the CBF, but is still limited to linear (first or second order) dynamics and struggles to extend to nonlinear motions.

Instead of approximating the observation dynamics, \cite{Harms24} proposes treating observations as parameters of safe sets in a switching approach, relying on the principle that a state safe under one observation remains safe over time, and provides a constructive approach of the LiDAR-based CBF (for an input-constrained second-order linear system).
Other works approximate a CBF from obstacle-centered SDFs~\cite{Long21} or via a GP~\cite{Keyumarsi24}, though both are restricted to first-order dynamics.

In general, these techniques struggle to faithfully encode observation-based safe sets for systems with complex (\ie, nonlinear or high-order) dynamics.
Our previous work~\cite{Jacquet24} proposes an alternative that consists of using the NN to encode a simpler, position-only representation in the form of an occupancy map, while the dynamics are handled by an NMPC.
However, the occupancy map gradient is nearly zero everywhere except at obstacle boundaries, limiting expressivity
Our proposed method overcomes this by encoding observations as an SDF.

\subsection{Implicit Distance Fields}

In computer vision, NNs are commonly used for implicit representation of surfaces or volumes, \ie, as the $0$-levelset of a scalar field.
Indeed, Coordinate-based networks~\cite{Tancik20} take spatial coordinates as input and output the corresponding Boolean occupancy~\cite{Chen19, Mescheder19} or SDF values~\cite{Park19}, offering continuous, compact representations encoded in network weights or latent spaces.
These methods have been applied to shape rendering~\cite{Wang21, Azinovic22, Wang22} and completion~\cite{Dai17, Stutz18}.

In robotics, SDFs are often organized as Euclidean Signed Distance Functions (ESDFs), assigning voxel-wise distances to the nearest obstacle.
However, obtaining an SDF from sensor data, online, can be a challenging task.
Classical approaches~\cite{Newcombe11, Oleynikova17} often construct these incrementally from Truncated Signed Distance Functions (TSDFs), which store truncated distances along sensor rays.
In~\cite{Han19} an efficient alternative using occupancy grids with fast indexing is proposed.

Consequently, neural SDFs have gained traction for mapping and localization.
Neural SLAM frameworks have been introduced in~\cite{Sucar21,Zhu22, Wang23}, where an SDF is encoded in an Multi-Layer Perceptron (MLP) and learned online from RGBD images.
Other approaches build SDFs online using TSDFs-inspired schemes~\cite{Yan21, Ortiz22}, with extensions for object structure~\cite{Jang24} and hierarchical representation~\cite{Vasilopoulos24}.
A comprehensive mapping framework with global consistency is proposed in~\cite{Pan24}.
Separately, a relevant, non-neural framework uses GP-based SDF for non-neural mapping, odometry, and planning~\cite{Wu23}.

Aside from mapping and navigation, implicit SDFs are also employed for manipulation,
\eg, as observations to an RL agent for grasping in~\cite{Mosbach22}.
Differently, in~\cite{Lee24} a network is trained to predict SDF of the swept volume of a manipulator arm, to ensure operator safety.
A non-neural example is provided in~\cite{Maric24}, where the SDF is approximated using piecewise polynomial SDFs with enforced $\mathcal{C}^1$ continuity.
While relying on few parameters compared to NNs, the scalability to large, complex scenes is questionable, and the construction time remains prohibitive for real-time applications.

\section{Problem Statement}\label{sec:problem}

We consider the problem of collision-free navigation in an unknown, static environment by exploiting the immediate high-resolution exteroceptive range observations.
We assume a reference velocity to be available, which is followed when possible while remaining collision-free.
The AR is assumed to be a rigid body,
whose state vector is denoted by $\vect{x} \in \mathcal{X}$
and the input vector is written as $\vect{u} \in \mathcal{U}$.
We assume that all collision constraints can be expressed as constraints on the position of the robot alone,
and that the system state $\vect{x}$ contains position information, denoted $\vect{p}\in\R^3$.

The system state evolves according to
\begin{equation}
    \dot{\vect{x}} = \vect{f}(\vect{x}, \vect{u}),
    \label{eq:pb:dyn}%
\end{equation}
where $\vect{f}$ denotes the nonlinear AR dynamics (detailed, \eg, in~\cite{Lee10}), which is recalled in \sect{sec:mpc:model}.
The AR also integrates a constrained-FoV range sensor that periodically produces observations $\vect{o} \in \mathcal{O}$ of a body-centric 3D volume defined by the sensor frustum, denoted $\mathbb{F}\subset\R^3$.
The observation $\vect{o}$ is a function of the (potentially uncertain) state vector $\vect{x}$ and the unknown environment. 
In turn, this implies that the mapping function $\xi$ between the state and the observation $\vect{o} = \xi (\vect{x})$ is unknown.

We specifically tailor the scope of this work to \mbox{$0$-memory} navigation,
\ie, at a given time $t$, the set $\mathcal{X}_\text{free}(t)$ of states which are not in collision is a function of $\vect{o}$.
Because of the constrained FoV, such a set can strictly be defined only within the visible volume $\mathbb{F}(t)$.
Accordingly, we define the mapping \mbox{$h:~\R^3\times\mathcal{O}\rightarrow\R$} as
\begin{equation}
    \mathcal{X}_\text{free}(t) =
        \{ \vect{x} \in \mathcal{X}~|~\vect{p}\in\mathbb{F}(t),
        ~h(\vect{p},\vect{o}) \ge 0 \}.
\end{equation}

This mapping $h$ defines the position constraint that must be satisfied to ensure collision avoidance.
However, the time derivative of $h$ cannot be obtained, as $\xi$ is unknown and in general discontinuous.
Instead, we resort to a switching update approach, as in~\cite{Harms24}, that avoids an explicit approximation of the observation dynamics.
At each step $k$, the discrete $\vect{o}_k$ defines a parametric mapping \mbox{$h_{\vect{o}_k}:~\R^3\rightarrow\R$},
which in turn defines (together with the corresponding frustum $\mathbb{F}_k$) the set $\mathcal{X}_{\text{free},k}$ of reachable positions w.r.t. the observation frame.

Our proposed method for tackling this problem is presented hereafter.
In \sect{sec:nn}, we present an approach for approximating $h_{\vect{o}}$,
while in \sect{sec:mpc} we present the actual system model and derive the subsequent definition of the optimal control problem.

\section{Learning the Environment Representation}\label{sec:nn}

Given the most recent observation $\vect{o}$,
we want to explicitly define the mapping $h_{\vect{o}}$ that defines the set of free positions within the sensor frustum $\mathbb{F}$.
However, the very high dimensionality of the observations (typically, $\approx1\mathrm{e}5$ pixels or $3$D points) prevents naive implementations that would directly transpose each pixel to individual constraints.
It is therefore necessary to use an intermediate volumetric representation of the visible environment.
However, classical approaches for computing said representations (\eg, raycasting or k-d tree search) suffer limitations such as
a)~slow computation,
b)~algorithmic computing, or
c)~discontinuous representations,
which in turn prevents their embedding in gradient-based optimal controllers.

The core of the proposed approach is to leverage deep learning to encode the sensor observation into a single volumetric constraint, using a neural implicit SDF constructed from a single depth image.

\subsection{SDF from observations}\label{sec:nn:sdf}

\begin{figure}
    \includegraphics[width=\linewidth]{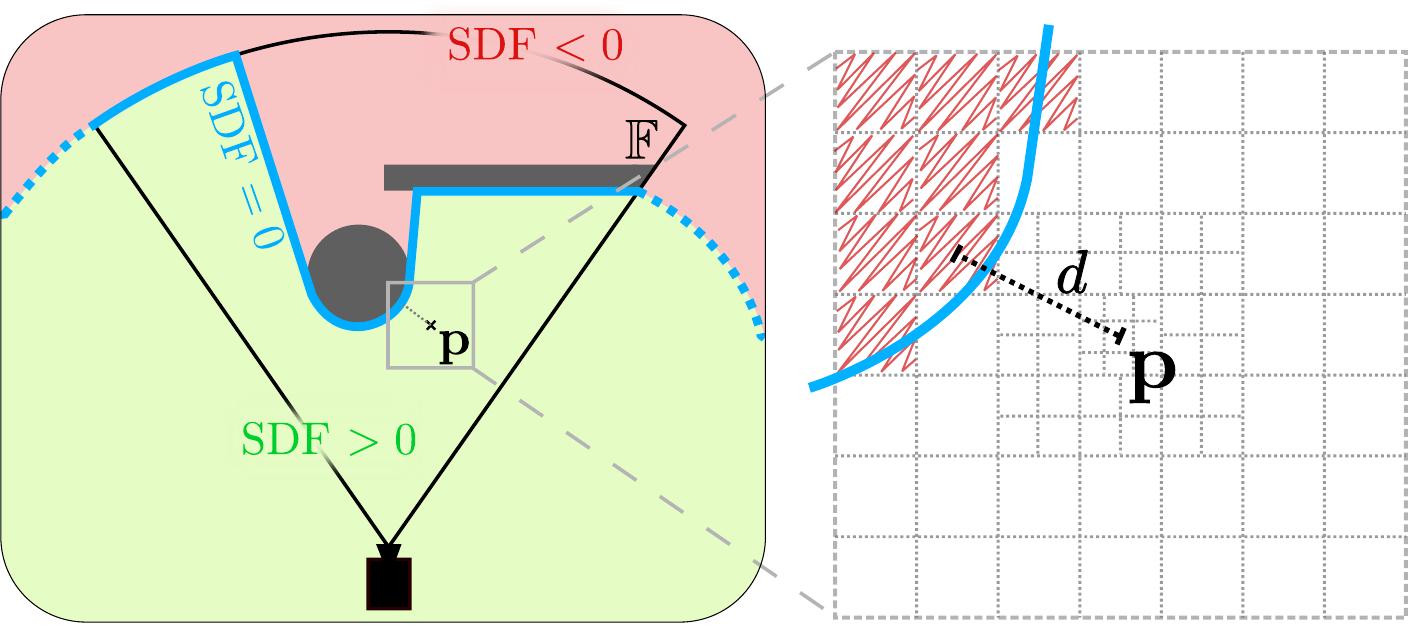}
    \caption{A 2D visualization of the distance transform (left) with two visible obstacles (gray).
        The blue line marks the $0$-level set of the SDF,
        and its dashed part illustrates its heuristic extension beyond the FoV $\mathbb{F}$ for training purposes.
        The zoom-in (right) illustrates the grid-based approximation of the distance transform for $\vect{p}$ by the distance $d$ between the central cell and the closest cell with a different occupancy (in red).}
    \label{fig:nn:sdf}
\end{figure}

We define the distance field as the spatial distance function to the surface of visible obstacles.
Specifically, it must have negative values for non-visible space, and positive values for visible space, as shown in Figure~\ref{fig:nn:sdf}.
Such a function is called a distance transform~\cite{Maurer03}.
As commonly done for neural SDFs~\cite{Park19}, with inherently limited representation capabilities, the SDF is truncated beyond some user-defined distance,
denoted~$T_\text{SDF}$.
For precise surface encoding, $T_\text{SDF}$ is typically a few centimeters.
For navigation, however, the obstacles are represented at a larger scale and we set $T_\text{SDF} = \SI{1}{m}$.

However, this distance transform is defined on \mbox{$\mathbb{F} \subset \R^3$}.
To ensure proper boundary conditioning during training, its definition is extended to $\R^3$ by assigning SDF values based on the spherical projection onto $\mathbb{F}$ (see \fig{fig:nn:sdf}).
This entails that an additional constraint must enforce the queried positions to be within $\mathbb{F}$, where the SDF truly relates to the observation.

\begin{figure*}[t]
    \centering
    \includegraphics[width=\linewidth]{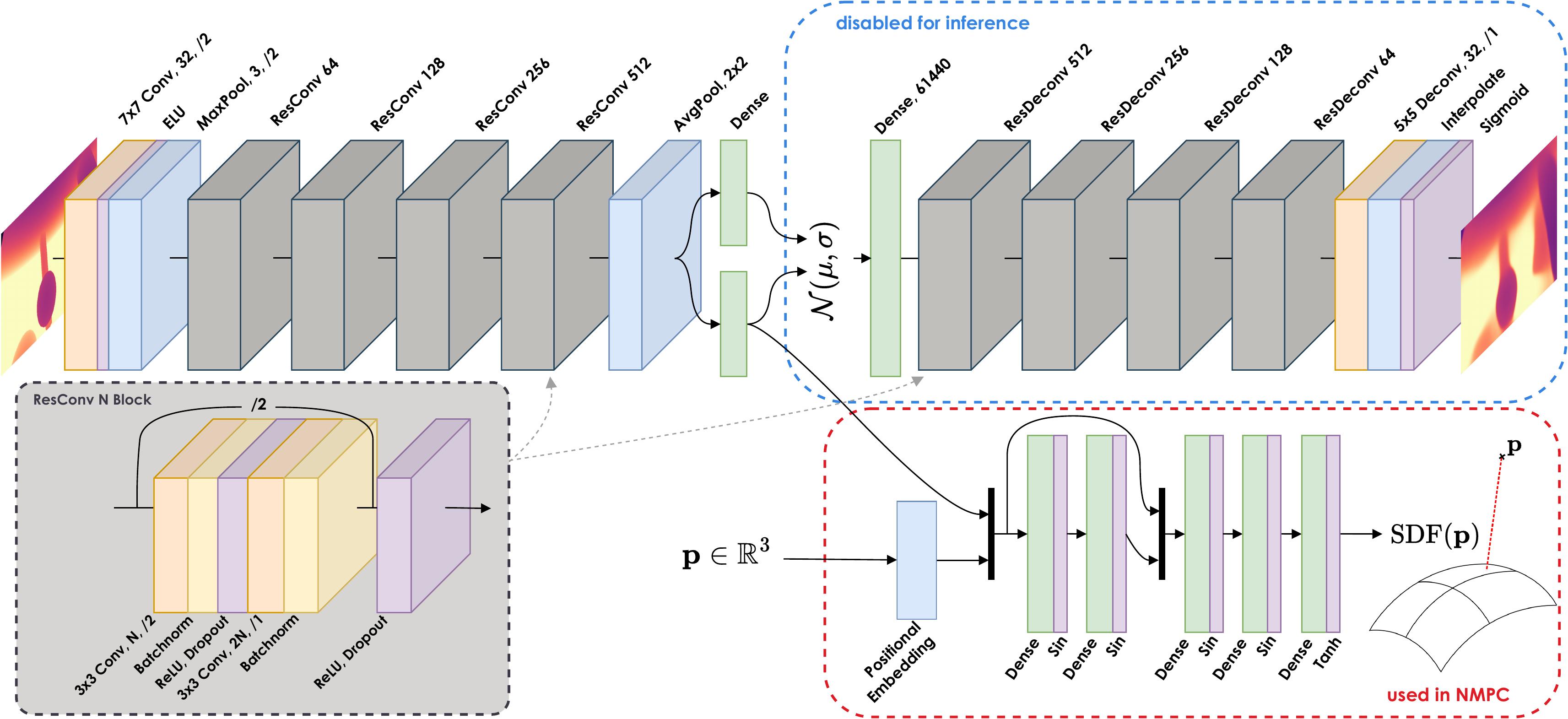}
    \caption{Architecture of the neural networks. Convolution and deconvolution layers are pictured in orange, linear layers in green, activations in purple, and pooling layers in blue.
    The encoder computes a Gaussian latent representation of the image, parametrized by its mean and std.
    The latent is sampled and decoded to reconstruct the input image.
    The decoder (blue rectangle) is used for training the VAE but is disabled for inference.
    The Residual block for a given size $N$ is pictured in the gray rectangle and instantiated in the network structure.
    Residual Deconvolution blocks follow the exact same structure.}
    \label{fig:nn:architecture}
\end{figure*}

Although there exist efficient parallelized algorithms for computing such a transform~\cite{Criminisi08, Cao10}, those are tailored for computing the full distance transform over a given grid.
This is typically not well suited for sampling-based training.
Instead, we utilize an approximation algorithm to generate the target values $\text{SDF}(\vect{p})$, for each sampled 3D position $\vect{p}$, given the observation.
This algorithm, illustrated in Figure~\ref{fig:nn:sdf}, consists of fitting a 3D grid of adaptive resolution on the sampled point, locating the closest cell where the occupancy changes.
The grid is dense around the center (thus enforcing higher precision close to surfaces)
but gets increasingly sparser for achieving efficient online computation.
The algorithm seamlessly provides the direction of the spatial gradient,
whose norm satisfies an eikonal equation

\begin{equation}
    \norm{\frac{\partial \text{SDF}(\vect{p})}{\partial \vect{p}}} =
    \begin{cases}
        1 \quad \text{if~} \norm{\text{SDF}(\vect{p})} < T_\text{SDF} \\
        0 \quad \text{otherwise}
    \end{cases}.
\label{eq:nn:eikonal}%
\end{equation}

We note that the precision of the approximation algorithm is therefore lower bounded by the thinnest resolution of the grid.
This algorithm can be tensorized for online training data generation.
It relies on a routine for evaluating the occupancy of arbitrary points given the input image,
which can be efficiently implemented on GPU kernels via backprojection onto the pixel grid, \eg, using Warp~\cite{WarpLib}.

\subsection{NN architecture}\label{sec:nn:arch}

We use a two-step process to handle observations.
First, a Convolutional Neural Network (CNN) encodes the observation into a compact representation.
This latent vector is then passed through a MLP for further processing.,
exploiting the spatial correlation properties of convolutions.

The two-step approach allows for
a)~avoid redundant computation when evaluating the SDF at multiple points given the same observation,
b)~leverage spatial patterns in the input via the CNN, while keeping the MLP (which parametrizes the collision constraint) lightweight, and finally
c)~improve robustness to the noise and systematic errors in the observations~\cite{Kulkarni23a} (\eg, stereo shadow and invalid LiDAR rays).

We use a $\beta$-VAE~\cite{Higgins17} architecture with a ResNet-10 network as the encoder.
It incorporates ReLU activations, batch normalizations, and dropout regularization for improved generalizability.
The final convolution volume is passed through an average pooling, before flattening and processing through an Fully Connected (FC) layer for computing the mean ($\vecg{\mu}$) and standard deviation (std) ($\vecg{\sigma}$) parametrization of the Gaussian space.
Accordingly, we use a deconvolution network based on ResNet-10 as the decoder.

The latent encoding $\vect{z}$ is processed using a coordinate-based MLP~\cite{Tancik20} that approximates the distance transform for the corresponding observation.

Similar to~\cite{Ortiz22}, positional embedding is first applied to the 3D point $\vect{p}$,
mapping it to a high-dimensional space using periodic activation functions.
This sinusoidal mapping of input coordinates allows MLPs to better represent higher frequency content~\cite{Tancik20}.
We use the off-axis positional embedding $\vecg{\gamma}$~\cite{Barron22}, defined as

\begin{equation}
    \vecg{\gamma}(\vect{p}) = \bmat{\vect{p} \\ \sin(2^0\mat{A}\vect{p}) \\ \cos(2^0\mat{A}\vect{p}) \\ \vdots \\ \sin(2^{L-1}\mat{A}\vect{p}) \\ \cos(2^{L-1}\mat{A}\vect{p})} \in \R^{2L + 3},
\end{equation}
where $L$ is the number of frequency bands, and the rows of $\mat{A}\in\R^{12\times3}$ are vertices of a unit-norm icosahedron.

This embedding is concatenated to $\vect{z}$, and processed via an MLP with $4$ hidden layers and using sine activations~\cite{Sitzmann20}.
The input to the MLP is also feed-forwarded to the third hidden layer.
The MLP also uses dropout regularization.
Various sizes for the hidden layers are evaluated in Sections~\ref{sec:valid:sdf} and~\ref{sec:valid:ablation}.

\begin{table}[t]
\centering
    \small
\begin{tabular}{c|cccc}
    \toprule
        \multirow{2}{*}{}
            & \multicolumn{2}{c}{\textbf{Depth camera}} & \multicolumn{2}{c}{\textbf{LiDAR}} \\
            & train            & test           & train        & test       \\
    \midrule
        Real       & 15165     & 1685           & 18682        & 2076       \\
        Gazebo     & 48039     & 5338           & 23707        & 2634       \\
        Aerial Gym & 91697     & 10189          & 89673        & 9964       \\
    \bottomrule
\end{tabular}
\normalsize

    \caption{Number of images used in the training and testing datasets.
        The sensors for each modality are an Intel D455 depth camera and an Ouster OS1 LiDAR.}
\label{tab:nn:training_dataset}
\end{table}

We note that the VAE operates directly on pixels,
and is therefore agnostic of the sensor FoV.
However, because of the convolution operations, the network is trained for a specific input image ratio,
which tends to differ between depth cameras (\eg, $16:9$) and LiDAR range images (\eg, $4:1$).

The SDF MLP operates on spatial data and implicitly encodes the intrinsic projection matrix, and is therefore trained for a specific sensor.

\subsection{Training procedure}\label{sec:nn:train}

The VAE and MLP are trained sequentially.

\subsubsection{VAE Training}\label{sec:nn:train:vae}

First, the VAE is trained using a dataset including images from real sensors,
as well as simulated images collected from Gazebo, and from Aerial Gym~\cite{Kulkarni25} based on Isaac Sim and using Warp rendering.
The first two sets consist of indoor and outdoor scenes,
while the latter consists of geometric shapes of arbitrary positions and orientations.
The number of images for each set is reported in \tab{tab:nn:training_dataset}.
The training set is randomly divided into training and validation subsets, using an 85\%-15\% ratio.
The input images are cropped up to a maximum range, and normalized to $\interv{0}{1}$.
The maximum range $d_\text{max}$ describes the range of environmental information encoded by the network.
In practice, we set $d_\text{max} = \SI{5}{m}$, which provides a sufficient horizon for effectively managing local collision avoidance.

We follow the training procedure in~\cite{Kulkarni23a},
\ie, no reconstruction loss is computed for the invalid pixels caused by stereo shadow or obstructions.

Because the encoder is used to prevent collisions, all obstacles must be reconstructed despite the high compression rate, in particular those close to the robot.
We therefore propose to bias the VAE to reconstruct close obstacles with higher accuracy.
This is achieved by scaling the loss function according to the distance value of the target pixel.
The scaling factor is computed such that it equals $1$ for $0$-valued pixels, and $w\in\intervexcl{0}{1}$ for $1$-valued pixels, where $w$ is a tunable parameter.
The intermediate values are obtained using a quadratic interpolation with a horizontal tangent at $0$.

The pixel-wise MSE reconstruction loss function of the VAE for an observation $\vect{o}$ is hence defined as
\begin{equation}
    \mathcal{L}_\text{MSE}(\vect{o}) = \sum_{\vect{o}_{ij} \neq 0}
        (\vect{o}_{ij}^2 (w - 1) + 1) (\text{VAE}(\vect{o})_{ij} - \vect{o}_{ij})^2,
\end{equation}
where the subscript $\bullet_{ij}$ refers to the pixel on the i\ith line and j\ith column.

Additionally, a $\beta$-scaled Kullback-Leibler Divergence (KLD) metric $\mathcal{L}_\text{KL}$ is used to enforce that the latent vector $\vect{z}$ follows a proper normal distribution fitting the true posterior,
as commonly done for variational encoders.
It is defined as
\begin{equation}
    \mathcal{L}_\text{KL} = -\frac{\beta_{\text{norm}}}{2}(1+\log(\vecg{\sigma}^2)-\vecg{\mu}^2-\vecg{\sigma}^2),
\end{equation}
where the $\vecg{\mu}$ and $\vecg{\sigma}$ are the predicted mean and std of the latent distribution,
and $\beta_{\text{norm}}$ is a scaling factor computed from the latent and image sizes, according to~\cite{Higgins17}.

\subsubsection{SDF Training}\label{sec:nn:train:sdf}

Then, the weights of the encoder are frozen, and the MLP is trained to approximate the SDF.
The training is also fully supervised, using regression of the target SDF value and gradient at sampled 3D points.
Target values for sampled points are computed following the procedure described in Section~\ref{sec:nn:sdf}.
The training is performed using the same dataset as the VAE.
However, because the SDF ground truth approximation relies on backprojection onto the image plane, the image must contain only valid pixels.
The images from the real sensors are therefore not used for training the MLP.
The sampling process is such that points are sampled
a)~within $\mathbb{F}$,
b)~in a ball centered on $O\us{S}$ (to enforce accuracy close to the origin),
c)~close to obstacles (for increased surface accuracy), and
d)~in a ball larger that $\mathbb{F}$ for boundary conditioning.
We use the sampling ratios $0.4$, $0.35$, $0.2$ and $0.05$, respectively.

The loss function is defined as the weighted sum of two MSE losses, for both the target SDF value and its gradient.

We note that the MSE loss on the unit norm gradient includes the satisfaction of the eikonal property \eqref{eq:nn:eikonal}.
We report, in \appen{app:training}, the training losses and metrics for both the VAE and SDF networks, which are used in the remainder of the paper.

\section{Neural NMPC}\label{sec:mpc}

\subsection{Mathematical Notations and Coordinate Frames}

We denote a coordinate frame $\bullet$ as $\frameV{\bullet}$,
its' origin as $O_\bullet$ and canonical base $(\vect{x}\us{\bullet}, \vect{y}\us{\bullet}, \vect{z}\us{\bullet})$.
We denote a generic world inertial frame with $\frameV{W}$.
The position and orientation of the AR are represented using the body-aligned frame $\frameV{B}$ attached to the geometric center of the robot $O\us{B}$.
Exteroceptive measurements made by the body-mounted range sensor are described in a sensor frame $\frameV{S}$.
Further, we define the vehicle frame~$\frameV{V}$, such that $O\us{V} = O\us{B}$, $\vect{z}\us{V} = \vect{z}\us{W}$, and which is yaw-aligned with $\frameV{B}$.

\begin{figure}[t]
    \centering
    \includegraphics[width=\columnwidth]{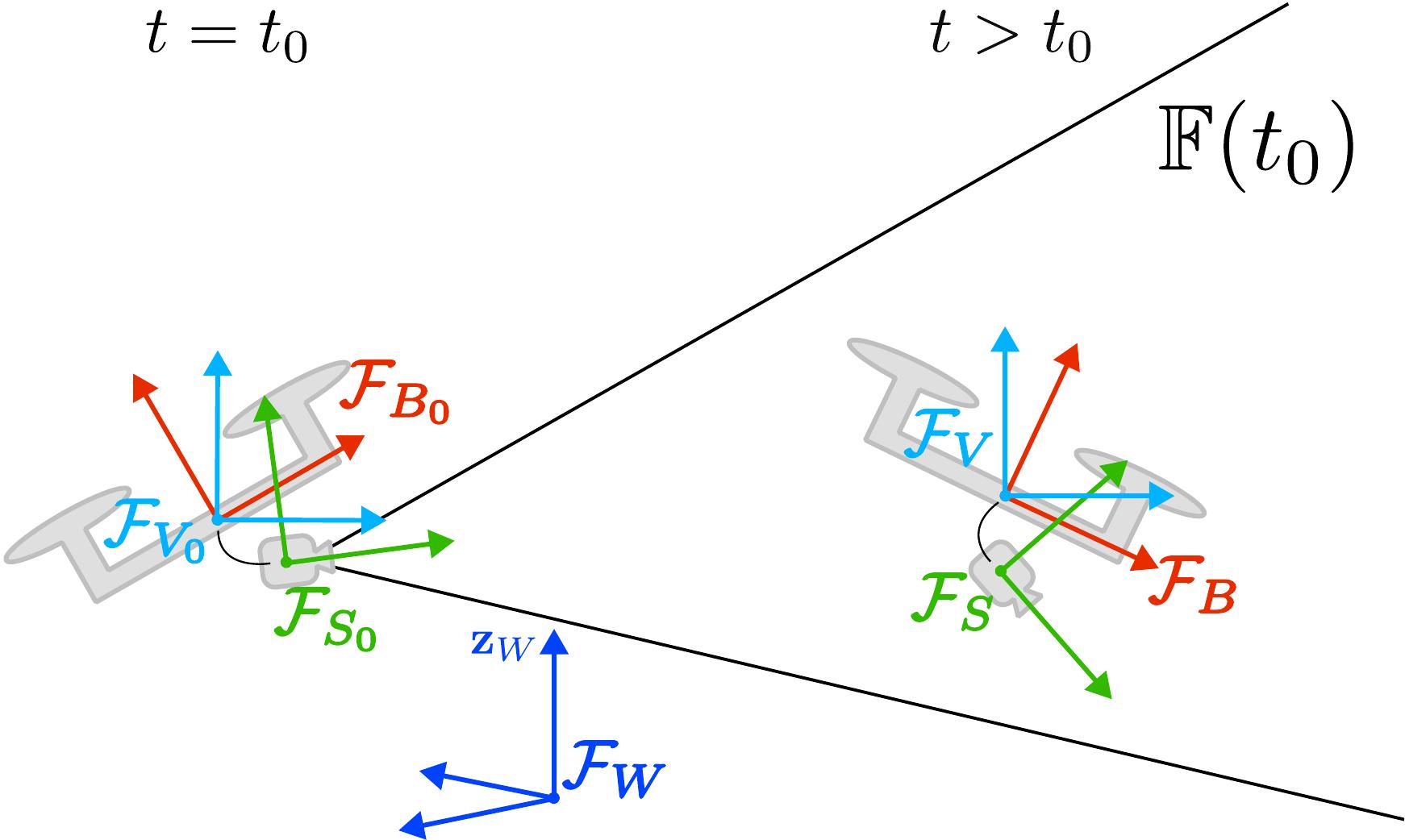}
    \caption{Planar representation of the relevant 3D frames used in the paper,
        including the inertial frames at time $t_0$ where the sensor observation is captured (left), and the sensor frustum $\mathbb{F}$.
        }
    \label{fig:mpc:frames}
\end{figure}

Lastly, we consider the inertial frames $\frameV{B_0}$, $\frameV{V_0}$, $\frameV{S_0}$, which at time $t=t_0$ coincide with $\frameV{B}$, $\frameV{V}$ and $\frameV{S}$, respectively.
A visual summary of the relevant frames is provided in \fig{fig:mpc:frames}.

The relative position of $\frameV{a}$ w.r.t. $\frameV{b}$, expressed in $\frameV{b}$, is denoted by $\ts{b}\vect{p}\us{a}$.
Similarly, the orientation of $\frameV{a}$ w.r.t. $\frameV{b}$ expressed in $\frameV{b}$ is denoted by $\ts{b}\mat{R}\us{a}$.
The unit quaternion representation of the rotation $\ts{b}\mat{R}\us{a}$ is denoted $\ts{b}\vect{q}\us{a}$,
and its Euler angle representation is $\ts{b}\vecg{\eta}\us{a} = \bmat{\phi~\theta~\psi}$, following the \textsc{zyx} Tait–Bryan convention.

Lower and upper bounds on any variable are respectively denoted with $\underline{\bullet}$ and $\overline{\bullet}$, and $\otimes$ denotes the Hamilton product of two quaternions.

\subsection{System Modeling}\label{sec:mpc:model}

The AR considered hereafter is a standard co-planar multirotor.
It is modeled as a rigid body of mass $m$
centered in $O\us{B}$, and actuated by $4$ or more co-planar propellers.
We assume that the robot is fully enclosed in a sphere of radius $r$.

We denote $t_0$ as the time at which the most recent observation is acquired.
We chose to represent the system position and orientation w.r.t. the inertial, switching frame $\frameV{V_0}$,
that is, the vehicle frame at $t_0$.
The heading with respect to $\frameV{V_0}$ is defined in quaternion form, denoted $\ts{V_0}\vect{q}\us{V}$.
It has $2$ non-zero components (the scalar and $z$ ones), respectively denoted $\ts{V_0}q\us{w}$ and $\ts{V_0}q\us{z}$.

Therefore, the system state is described, dropping the subscript $\bullet\us{B}$ for legibility, by the vector

\begin{equation}
    \vect{x} = \bmat{\ts{V_0}\vect{p}\transp, ~ \ts{V_0}q\us{w}, ~ \ts{V_0}q\us{z},  ~ \ts{V_0}\vect{v}\transp}\transp \in\R^{8},
\label{eq:mpc:x}%
\end{equation}
where $\ts{V_0}\vect{v}$ is the velocity of $O\us{B}$, expressed in $\frameV{V_0}$.

Accordingly, we select the system input variables in order to control the $3$D thrust (magnitude and orientation through roll and pitch) and yaw rate, as typically done for co-planar multirotors~\cite{Furrer16}.
The system input vector is therefore

\begin{equation}
    \vect{u} = \bmat{T, ~ \ts{V}\phi, ~ \ts{V}\theta, ~ \ts{B}\omega\us{z}}\transp \in\R^{4},
\label{eq:mpc:u}%
\end{equation}
where $T$ is the collective thrust of the actuators along $\vect{z}\us{B}$,
$\ts{V}\phi$ and $\ts{V}\theta$ are the roll and pitch angles of $\ts{V}\mat{R}\us{B}$,
and $\ts{B}\omega\us{z}$ is the angular rate around $\vect{z}\us{B}$, expressed in $\frameV{B}$.
It is assumed that lower and upper bounds on $\vect{u}$ are known, or obtained through an identification campaign.

The system dynamics \eqref{eq:pb:dyn} considered in the problem formulation are therefore instantiated as

\begin{subequations}
    \begin{align}
        \ts{V_0}\dot{\vect{p}} &= \ts{V_0}\mat{R}\us{B} \ts{V_0}\vect{v}, \\
        \bmat{\dot{\ts{V_0}q\us{w}}\\0\\0\\\dot{\ts{V_0}q\us{z}}} &= \frac{1}{2}~\bmat{\ts{V_0}q\us{w}\\0\\0\\\ts{V_0}q\us{z}}\otimes\bmat{0\\0\\0\\\ts{B}\omega\us{z}}, \\
        \ts{V_0}\dot{\vect{v}} &= \frac{T}{m}\;\ts{V_0}\mat{R}\us{B}\vect{z}\us{B} - g\vect{z}\us{V},\label{eq:mpc:dyn:zeta}%
    \end{align}
\label{eq:mpc:dyn}%
\end{subequations}
where $\ts{V_0}\mat{R}\us{B} = \ts{V_0}\mat{R}\us{V} \ts{V}\mat{R}\us{B}$ is the composition of the rotations described respectively by $\ts{V_0}\vect{q}\us{V}$ and $(\ts{V}\phi, \ts{V}\theta)$,
and $g$ is the magnitude of the gravity vector.

We further motivate this choice of dynamics representation in~\appen{app:dyn}.

The onboard range sensor, whose principal axis is $\vect{x}\us{S}$,
is rigidly attached such that $\ts{B}\vect{p}\us{S}$ and $\ts{B}\mat{R}\us{S}$ are constant and known.
Its FoV $\mathbb{F}$ is described by the halved vertical and horizontal angular apertures,
respectively $\alpha\us{V}$ and $\alpha\us{H}$,
as well as a maximum distance $d_{\text{max}}$.
In practice, we set $d_{\text{max}} = \SI{5}{m}$ and match the maximum range handled by the VAE and SDF networks.

\subsection{Local Navigation Neural MPC}\label{sec:mpc:navmpc}

\subsubsection{Obstacle Avoidance Constraint}\label{sec:mpc:colav_const}

We assume that the observation $\vect{o}$ is encoded into a latent representation $\vect{z}$ through the VAE (which is a convenience wording shortcut referring to taking the mean of the latent distribution through the encoder part of the VAE).
The neural approximation of the SDF defined in Section~\ref{sec:nn:sdf} defines the function

\begin{equation}
    \begin{split}
        \R^3 &\rightarrow \R \\
        \vect{p} &\mapsto \text{SDF}_{\vecg{\theta},\vect{z}}(\vect{p})~,
    \end{split}
\end{equation}
which is parametrized by~$\vect{z}$ and the MLP weights, denoted as~$\vecg{\theta}$.
Ensuring that the motion of the AR remains collision-free is achieved by constraining the SDF value of positions of the open-loop trajectory w.r.t. the observation $\vect{o}$.
Recalling that $\vect{o}$, acquired at $t=t_0$, describes a position constraint w.r.t. $\frameV{S_0}$, we write the position of $O_B$ in $\frameV{S_0}$
as a closed-form function of the NMPC state and
parametrized by the roll and pitch angles at time $t_0$,
as

\begin{equation}
    \ts{S_0}\vect{p}\us{B} =
        \ts{B}\mat{R}\us{S}\transp
        \left( \ts{V_0}\mat{R}\us{B_0}\transp \ts{V_0}\vect{p}\us{B} - \ts{B}\vect{p}\us{S} \right).
\end{equation}

Then, we impose on the NMPC the constraint
\begin{equation}
    \text{SDF}_{\vecg{\theta},\vect{z}}(\ts{S_0}\vect{p}\us{B}) \ge r + \epsilon,
\label{eq:mpc:const_sdf}%
\end{equation}
where $\epsilon > 0$ is a user-defined safety margin added to the robot-enclosing radius $r > 0$.

\subsubsection{Field of View Constraints}\label{sec:mpc:navmpc:fov_const}

Because the collision-free set is defined in the sensor frustum $\mathbb{F}(t_0)$,
the predicted motion must evolve within this volume.
This results in a set of two additional constraints,
respectively relative to the horizontal and vertical translations of the sensor $O_S$ w.r.t. to $\frameV{S_0}$.

We denote the transformations from Euclidean coordinates to the azimuth and elevation angles as $\mathcal{S}_\text{azimuth}\colon \R^3 \rightarrow \R$ and $\mathcal{S}_\text{elevation}\colon \R^3 \rightarrow \R$, respectively, which are computed as
\begin{equation}
\begin{aligned}
    \mathcal{S}_\text{azimuth}(\smat{x\\y\\z}) &= \mathrm{atan}2 (y, x),\\
    \mathcal{S}_\text{elevation}(\smat{x\\y\\z}) &= \mathrm{atan}2(z, \sqrt{x^2 + y^2}).
\label{eq:mpc:spherical_coords}%
\end{aligned}
\end{equation}

The resulting constraints are then written as
\begin{equation}
\begin{aligned}
    -\alpha\us{H} \le \mathcal{S}_\text{azimuth}(\ts{S_0}\vect{p}\us{S}) \le \alpha\us{H},\\
    -\alpha\us{V} \le \mathcal{S}_\text{elevation}(\ts{S_0}\vect{p}\us{S}) \le \alpha\us{V}.
\label{eq:mpc:const_fov}%
\end{aligned}
\end{equation}
We note that an additional maximum range constraint would be redundant,
as it is already encoded in \eqref{eq:mpc:const_sdf} that every position past $d_{\text{max}}$ is unsafe.

\subsubsection{Objective Function for Velocity Tracking}\label{sec:mpc:navmpc:cost}

We assume that a reference velocity and a reference heading, both expressed w.r.t. $\frameV{V_0}$, are provided by any high-level planner,
respectively denoted $\ts{V_0}\vect{v}_\text{ref}$ and $\ts{V_0}\psi_\text{ref}$.

Following~\cite{Brescianini20}, we express the heading errors in quaternions.
Let $\ts{V_0}\vect{q}_\text{ref}$ be the quaternion representation of $\ts{V_0}\psi_\text{ref}$.
The orientation error $\vect{q}\us{e}$ is given by

\begin{equation}
    \vect{q}\us{e}
    = \bmat{q\us{e,w}\\0\\0\\q\us{e,z}}
    = \ts{V_0}\vect{q}_\text{ref} \otimes \bmat{\ts{V_0}q\us{w}\\0\\0\\\ts{V_0}q\us{z}}\inv,
\label{eq:mpc:q_error}
\end{equation}
and minimizing the heading error is therefore equivalent to minimizing $q\us{e,z}$.

Additionally, a control input objective penalizes the magnitude of the roll, pitch, and yaw rate,
as well as the deviation from $mg$ of the vertical projection of the thrust $T_z = T\cos(\ts{V}\phi)\cos(\ts{V}\theta)$.

The stage cost $\ell(\vect{x}, \vect{u})$ is thus written as

\begin{equation}
    \ell(\vect{x}, \vect{u}) =
    \norm{\bmat{
        q\us{e,z} \\
        \ts{V_0}\vect{v} - \ts{V_0}\vect{v}_\text{ref}
    }}^2_\mat{Q}
    + \norm{\bmat{
        T_z - mg \\
        \ts{V}\phi \\
        \ts{V}\theta \\
        \ts{B}\omega\us{z}
    }}^2_\mat{R},
    \label{eq:mpc:cost}
\end{equation}
where $\mat{Q}$ and $\mat{R}$ are tunable positive semidefinite and positive definite weight matrices, respectively,
and $\norm{\bullet}^2_\mat{M}$ is the weighted squared norm operator defined as

\begin{equation}
    \norm{\vect{a}}^2_\mat{M} = \vect{a}\transp \mat{M}\vect{a},
\label{eq:mpc:w_sq_norn}
\end{equation}
for an arbitrary vector $\vect{a}$ and diagonal weight matrix $\mat{M} \ge 0$.

\subsection{Theoretical Analysis}\label{sec:mpc:theory}

In this study, we construct a terminal constraint to ensure recursive feasibility of the control law.
Specifically, we enforce the terminal state to be such that there exists a sub-optimal terminal control policy that is recursively feasible under all constraints.

We then derive sufficient conditions on an appropriate terminal cost to ensure that the optimal cost becomes a non-increasing Lyapunov function under the sub-optimal terminal control policy.
This step is then used to quantify a region of local stability that scales with the choice of the free parameters in the terminal cost.

For the remainder of the analysis, we consider the evolution of the system dynamics under a fixed observation.
This means that we effectively neglect inaccuracies of the SDF approximation and the observation-dependent nature of $\mathcal{X}_\text{free}$ by assuming that $\frameV{S_0}$ and $\text{SDF}_{\vecg{\theta},\vect{z}}$ are not time-varying.
This partial approach does not directly prove the feasibility of the overall control problem under switching and noisy observations.
However, it ensures that the control law never enters unsafe states under the current observation.
Since the trajectory remains within the sensor frustum, this approach ensures that the (extended) open-loop trajectory is collision-free at all times.

The yaw dynamics are disregarded since they are decoupled from position dynamics for the considered AR.

\subsubsection{Recursive Feasibility}\label{sec:mpc:theory:recfeas}

For all terminal states $\vect{x}_N$, we define a corresponding set $\mathcal{X}_N(\vect{x}_N)$ as the set of all states reached by recursively applying a ``maximum braking to standstill'' policy.
This set is trivially forward invariant for such a policy and is bounded (both in velocity and position).
To make the braking policy recursively feasible, it remains to ensure that it satisfies the input constraints, and that all states in $\mathcal{X}_N(\vect{x}_N)$ satisfy the collision avoidance constraints, \ie, $\forall \vect{x} \in \mathcal{X}_N(\vect{x}_N)$,

\begin{subequations}
\begin{gather}
    \text{SDF}_{\vecg{\theta},\vect{z}}(\ts{S_0}\vect{p}) \ge r + \epsilon, \label{eq:mpc:rec_feas_cond:sdf} \\
    \ts{S_0}\vect{p} \in \mathbb{F}(t_0). \label{eq:mpc:rec_feas_cond:fov}
\end{gather}
\label{eq:mpc:rec_feas_cond}%
\end{subequations}

The maximum braking policy, denoted $\pi_\text{b}(\vect{x})$, is defined as the solution to the constrained optimization problem

\begin{subequations}
\begin{equation}
    \{\pi_\text{b}(\vect{x}), \lambda_b\} = \argmax_{T, \phi, \theta, \lambda}{\norm{\vect{a}_\text{b}(\vect{v})}}
\end{equation}%
\begin{align}
    s.t. ~~~
    & \vect{a}_\text{b}(\vect{v}) = \frac{T}{m}\ts{V_0}\mat{R}\us{B}\vect{z}\us{B} - g\vect{z}\us{V} \\
    & \vect{a}_\text{b}(\vect{v}) = -\lambda\vect{v} \label{eq:mpc:a_brake:line} \\
    & \lambda \ge 0 \label{eq:mpc:a_brake:overshoot} \\
    & 0 \le T \le \overline{T} \label{eq:mpc:a_brake:const_T} \\
    & \underline{\phi} \le \phi \le \overline{\phi} \label{eq:mpc:a_brake:const_roll} \\
    & \underline{\theta} \le \theta \le \overline{\theta} \label{eq:mpc:a_brake:const_pitch}
\end{align}
\label{eq:mpc:a_brake}%
\end{subequations}
where $\vect{a}_\text{b}(\vect{v})$ is the deceleration opposite to the velocity vector,
and Equations~\eqref{eq:mpc:a_brake:const_T},~\eqref{eq:mpc:a_brake:const_roll} and~\eqref{eq:mpc:a_brake:const_pitch} are the input constraints imposed on the system.
The policy ensures a deceleration along a straight line via Eq.~\eqref{eq:mpc:a_brake:line},
and the system position thus evolves along a straight line toward the hovering equilibrium.
Therefore, it holds that

\begin{equation}
    \forall \vect{x} \in \mathcal{X}_N, \norm{\vect{p} - \vect{p}_N} \le d_\text{b},
\end{equation}
where $d_\text{b}$ is the braking distance traveled while applying $\pi_\text{b}$ given by

\begin{equation}
    d_\text{b}(\vect{v}_N) = \frac{\vect{v}_N^2}{2~\norm{\vect{a}_\text{b}(\vect{v}_N)}},
\label{eq:mpc:d_brake}%
\end{equation}
since the system is subject to a constant deceleration $\vect{a}_\text{b}(\vect{v}_N)$.

The condition~\eqref{eq:mpc:rec_feas_cond:sdf} is thus enforced by the terminal constraint

\begin{equation}
    \text{SDF}_{\vecg{\theta},\vect{z}}(\ts{S_0}\vect{p}\us{B,N}) - d_\text{b} \ge r + \epsilon.
\label{eq:mpc:brake_constraint}%
\end{equation}

Because $\mathbb{F}(t_0)$ is convex, the second condition~\eqref{eq:mpc:rec_feas_cond:fov} is satisfied if

\begin{equation}
    \ts{S_0}\vect{p}\us{S,N} \in\mathbb{F}(t_0),\quad \ts{S_0}\vect{p}\us{S,E} \in\mathbb{F}(t_0),
\label{eq:mcp:term_fov_const}%
\end{equation}
where $\vect{p}\us{S,E}$ is the position of $O_S$ in the hovering equilibrium, obtained by translation of $d_\text{b}(\vect{v}_N)$ along the direction of $\vect{v}_N$.
This constraint is enforced as in Eq.~\eqref{eq:mpc:const_fov}.

To include these two terminal constraints in the Optimal Control Problem, we compute a closed-form approximation of the braking distance $d_\text{b}$.
We employ an i\ith degree \mbox{$3$-variate polynomial}, whose coefficients are obtained through a least-square fitting on the target values obtained through Equations~\eqref{eq:mpc:a_brake} and \eqref{eq:mpc:d_brake}.
\tab{tab:mpc:braking_dist_fit} reports the approximation errors for a given set of actuation constraints,
evaluating polynomials of various degrees to provide insights on the expected accuracy.
We note that directly approximating $d_\text{b}$ is easier than approximating $\vect{a}_\text{b}$ since the latter is discontinuous in $\vect{v} = 0$, while \eqn{eq:mpc:d_brake} can be trivially extended by continuity.
Hereafter, we assume that an approximation of $d_\text{b}$ of arbitrary precision is available.

\begin{table}[t]
\centering
    \small
\begin{tabular}{c|ccccc}
    \toprule
        {\textbf{\makecell{Degree of the \\ fitted polynomial}}}
        & 3
        & 4
        & 5
        & 6
        & 7 \\
    \midrule
        \textbf{\# of params.}
            & 20
            & 35
            & 56
            & 84
            & 120 \\
    \midrule
        \textbf{RMSE} [\si{cm}]
            & 4.0
            & 2.6
            & 2.4
            & 1.7
            & 1.6 \\
    \bottomrule
\end{tabular}
\normalsize

    \caption{Fitting error, in centimeters, of various polynomial approximations of $d_\text{b}(\vect{v})$ as computed through Equations~\eqref{eq:mpc:a_brake} and \eqref{eq:mpc:d_brake}.
        The results are evaluated on the ball $\norm{\vect{v}} \le 3$, with a resolution of $0.05$ (\ie, $\approx \SI{0.9}{M}$ points).}
\label{tab:mpc:braking_dist_fit}
\end{table}

\subsubsection{Stability}\label{sec:mpc:theory:stab}

The current problem setting does not allow the establishment of asymptotic stability under arbitrary reference velocities
due to the conflicting objectives in the stage cost (reference velocity tracking) and terminal cost and policy (coming to a stop).
Further, the collision constraint may prevent the AR from converging to the desired reference velocity.
An intuitive example of this is when a forward reference velocity is given while a wall is blocking the way; in this case, the constraint brings the system to a full stop.
As a result, asymptotic stability (to a zero-cost equilibrium) cannot be established in general.

We instead derive sufficient conditions such that the optimal cost of the controller remains non-increasing.
This is achieved by selecting an appropriate terminal cost $V(\vect{x}_N)$ ensuring that the optimal cost, denoted $J^*$ and introduced in section \ref{sec:mpc:nlp}, remains non-increasing under the sub-optimal
terminal braking policy~$\pi_\text{b}$.
This enforces the existence of a feasible solution at the next time step that bounds $J^*$.

We emphasize that while stability is established under certain assumptions on the reference velocity $\vect{v}_\text{ref}$, the region in which the cost function is guaranteed to be non-increasing can be made arbitrarily large by the choice of the terminal cost parameter $p$.
This is sufficient for the intended use case, as hard constraints are acting on the system state to ensure feasibility.

The Lyapunov stability condition on $J^*$ is written

\begin{equation}
    J^*(\vect{x}_{t+1}) - J^*(\vect{x}_t) \le 0.
\label{eq:mpc:stab_condition}%
\end{equation}
Expanding the two terms, this condition can be equivalently written as a condition on the terminal cost $V(\vect{x})$

\begin{equation}
    V(\vect{x}_{N+1}) - V(\vect{x}_N)
    \le - \ell(\vect{x}_N, \pi_\text{b}(\vect{x}_N)),
\label{eq:mpc:term_cost_cond}%
\end{equation}
where $\ell(\vect{x}_N, \pi_\text{b}(\vect{x}_N))$ is the stage cost, evaluated in $\vect{x}_N$ while applying the braking policy~\eqref{eq:mpc:a_brake}.

We define the terminal cost as

\begin{equation}
    V(\vect{x}) = p~\vect{v}\transp\vect{v},
\label{eq:mpc:term_cost}%
\end{equation}
where $p > 0$.
Specific conditions on $p$ ensuring that Eq.~\eqref{eq:mpc:stab_condition} holds
are provided in \appen{app:stab}, including a narrow case when the system terminal velocity vanishes to zero (as well as a strategy to overcome this limitation by using a reference governor).

\subsection{NonLinear Programming}\label{sec:mpc:nlp}

The discrete-time NonLinear Programming (NLP) over the receding horizon $T$,
sampled in $N$ shooting points, at a given instant $t$,
given a range image captured at $t_0 \le t$ and compressed into a latent vector $\vect{z}$ via a network parametrized by $\vecg{\theta}$,
is expressed as
\begin{subequations}
\begin{equation}
J^* \quad = \min_{\substack{\vect{x}_0\dots\vect{x}_N \\ \vect{u}_0\dots\vect{u}_{N-1}}}
    \sum_{k=0}^{N-1}
        \ell(\vect{x}_k,\vect{u}_k) + V(\vect{x}_N)
\label{eq:mpc:nlp:cost_function}%
\end{equation}%
\begin{align}
    s.t. ~~
    &\vect{x}_0 = \vect{x}(t) & \\
    &\vect{x}_{k+1} = \vect{f}(\vect{x}_k,\vect{u}_k),
        &{\scriptstyle k\in\{0, N-1\}} \\
    &\underline{\vect{v}} \le \vect{v}_k \le \overline{\vect{v}},
        &{\scriptstyle k\in\{0, N\}}\label{eq:mpc:nlp:v_constr} \\
    &\underline{\vect{u}} \le \vect{u}_k \le \overline{\vect{u}},
        &{\scriptstyle k\in\{0, N-1\}}\label{eq:mpc:nlp:u_constr} \\
    &r + \epsilon \le \text{SDF}_{\vecg{\theta},\vect{z}}(\ts{S_0}\vect{p}\us{B,k}),
        &{\scriptstyle k\in\{0, N-1\}}\label{eq:mpc:nlp:avoidance} \\
    &-\alpha\us{H} \le \mathcal{S}_\text{azimuth}(\ts{S_0}\vect{p}\us{S,k}) \le \alpha\us{H},
        &{\scriptstyle k\in\{0, N\}}\label{eq:mpc:nlp:hfov_constr} \\
    &-\alpha\us{V} \le \mathcal{S}_\text{elevation}(\ts{S_0}\vect{p}\us{S,k}) \le \alpha\us{V},
        &{\scriptstyle k\in\{0, N\}}\label{eq:mpc:nlp:vfov_constr} \\
    &r\le\text{SDF}_{\vecg{\theta},\vect{z}}(\ts{S_0}\vect{p}\us{B,N}) - d_\text{b}(\vect{v}_N),
        &\label{eq:mpc:nlp:rec_feas:sdf} \\
    &-\alpha\us{H} \le \mathcal{S}_\text{azimuth}(\ts{S_0}\vect{p}_E) \le \alpha\us{H},
        &\label{eq:mpc:nlp:rec_feas:hfov} \\
    &-\alpha\us{V} \le \mathcal{S}_\text{elevation}(\ts{S_0}\vect{p}_E) \le \alpha\us{V},
        &\label{eq:mpc:nlp:rec_feas:vfov}
\end{align}
\label{eq:mpc:nlp}%
\end{subequations}
where $\vect{x}(t)$ is the state estimate at $t$,
and $\vect{f}$ synthetically denotes the discretized dynamics defined in \eqn{eq:mpc:dyn}.
We note that the NLP could be solved without the terminal constraint and cost related to recursive feasibility and stability.

\subsection{Implementation Details}\label{sec:mpc:impl}

The above NLP is implemented in Python using Acados~\cite{AcadosLib} and Casadi~\cite{CasADiLib}.
The neural network is implemented with PyTorch,
and it is interfaced with the NMPC using L4Casadi~\cite{Salzmann23}.
The NLP is transformed into an Sequential QP solved with Interior Point Method~\cite{HPIPMLib} with a Real-Time Iteration (RTI) scheme.

We use Levenberg–Marquardt (LM) regularization to improve the stability of computing sensitivities.
We remark that this has a strong impact on the stability of the SQP solution under switching observations, and thus switching constraints.
This is a consequence of both using a neural constraint parametrized with a large number of neurons,
and of using the RTI suboptimal solving strategy, which relies on a reliable warm start from the previous solution.
We empirically set the LM regularization factor to~$10$.

The FoV and neural constraints~\eqref{eq:mpc:nlp:avoidance},~\eqref{eq:mpc:nlp:hfov_constr}, and ~\eqref{eq:mpc:nlp:vfov_constr} are implemented as slackened constraints.
This allows both to retain feasibility w.r.t. noisy feedback on the initial state $\vect{x}_0$,
and w.r.t. the switching observations and the corresponding SDF approximation errors.
We impose a $L1$-norm penalty with a weight of $20$ on both slack variables for the FoV constraints,
and both $L1$ and $L2$ penalties on the avoidance constraint, with respective weights $200$ and $20$.
The large penalty on the neural constraint aims to compensate for the potentially large magnitude of the velocity tracking error
when the environment imposes a full stop on the robot (\eg, in front of the wall with a large forward velocity reference).

The proposed method relies on only a few tunable hyperparameters.
Those are, namely, the MPC weight (including those for the slack variables) which are tuned as for any other controller; the length $T$ and sampling $N$ of the horizon; the LM regularization factor; the safety margin $\epsilon$; and the size of the SDF network, which is discussed in the next subsection.
We note that the safety margin $\epsilon$ is chosen to be at least greater than the neural SDF RMSE, evaluated in \sect{sec:valid:sdf}.
\tab{tab:mpc:params} presents the values for these parameters for each experimental subsection hereafter.
The LM regularization factor and slack variables weights detailed in this section are not repeated, as they are constant in all the simulations and experiments.

\begin{table}[t]
\centering
    \small
\begin{tabular}{c|ccccc}
    \toprule
        Section &
        diag($\mat{Q}$) &
        diag($\mat{R}$) &
        $T$ [s] &
        $N$ &
        $\epsilon$ [m] \\
    \midrule
        Section~\ref{sec:valid:ablation} &
        $\smat{20 \\ 5 \\ 5 \\ 5}$ &
        $\smat{0.04 \\ 50 \\ 50 \\ 5}$ &
        $1.5$ & 
        $20$ & 
        $0.1$ \\
    \midrule
        Section~\ref{sec:valid:comparison} &
        $\smat{20 \\ 5 \\ 5 \\ 2}$ &
        $\smat{0.04 \\ 25 \\ 25 \\ 5}$ &
        $1.5$ & 
        $20$ & 
        $0.2$ \\
    \midrule
        Section~\ref{sec:valid:drift_odom} &
        $\smat{20 \\ 5 \\ 5 \\ 2}$ &
        $\smat{0.04 \\ 25 \\ 25 \\ 5}$ &
        $1.5$ & 
        $20$ & 
        $0.2$ \\
    \midrule
        Section~\ref{sec:valid:xp} &
        $\smat{20 \\ 5 \\ 5 \\ 5}$ &
        $\smat{0.04 \\ 50 \\ 50 \\ 5}$ &
        $1.5$ & 
        $10$ & 
        $0.3$ \\
    \bottomrule
\end{tabular}
\normalsize

    \caption{SDF-NMPC parameters used in the experimental sections.}
    \label{tab:mpc:params}
    \vspace{-2ex}
\end{table}

\section{Validation}\label{sec:valid}

In this section, we evaluate the components of the proposed method.
First, the encoding capabilities of the VAE and SDF networks are evaluated.
Then, the neural controller is evaluated through randomized simulations to assess its sensitivity to the various parameters.
The proposed method is then compared against two state-of-the-art mapless navigation methods.
We perform an ablation study to quantify the resilience to odometry drift.
Finally, the controller is integrated into a real system and validated in hardware experiments.

In order to properly study the properties of the proposed NMPC method, we avoid relying on a map-informed high-level planner generating the velocity reference provided to the NMPC.
Instead, we consider a worst-case setting and implement a naive, obstacle-agnostic goal-seeking planner.
It provides a reference $\ts{V_0}\vect{v}_\text{ref}$ along the straight line to reach the given goal position.
The magnitude $v_\text{ref}$ of $\ts{V_0}\vect{v}_\text{ref}$ is a user-defined parameter.
In the laterally-constrained FoV case, the planner also provides a reference heading $\ts{V_0}\psi_\text{ref}$ such that the sensor aligns with the direction of motion.

\subsection{VAE Evaluation}\label{sec:valid:vae}

This subsection presents a qualitative and quantitative evaluation of the VAE reconstruction.
We chose a latent space dimension of $128$, which offers a good trade-off between quality and compression for the same input image size ($480\times270$ pixels)~\cite{Kulkarni23a}.
We compare the proposed distance-weighted (biased) VAE, introduced in Section~\ref{sec:nn:train}, with a standard (unbiased) VAE.
The tuning parameter is fixed at $w = 0.01$.
We also include a baseline compression method based on the Fast Fourier Transform (FFT), which retains the $64$ complex frequencies (i.e., $128$ real values) with the largest magnitudes.

Quantitative results on the test dataset are summarized in \tab{tab:valid:vae}.
These results show that the biased VAE shifts reconstruction error toward the background, yielding better reconstruction in the foreground \ie, for the obstacles of interest for navigation.
Overall, it outperforms both the unbiased VAE and the FFT-based reconstruction.
Reconstruction quality is generally lower for LiDAR images compared to depth images,
since the broader FoV perceives more objects and the resulting image has a higher spatial frequency.
Thus, more information is passed through the same latent bandwidth.

\begin{figure}[t]
    \centering
    \includegraphics[width=\columnwidth]{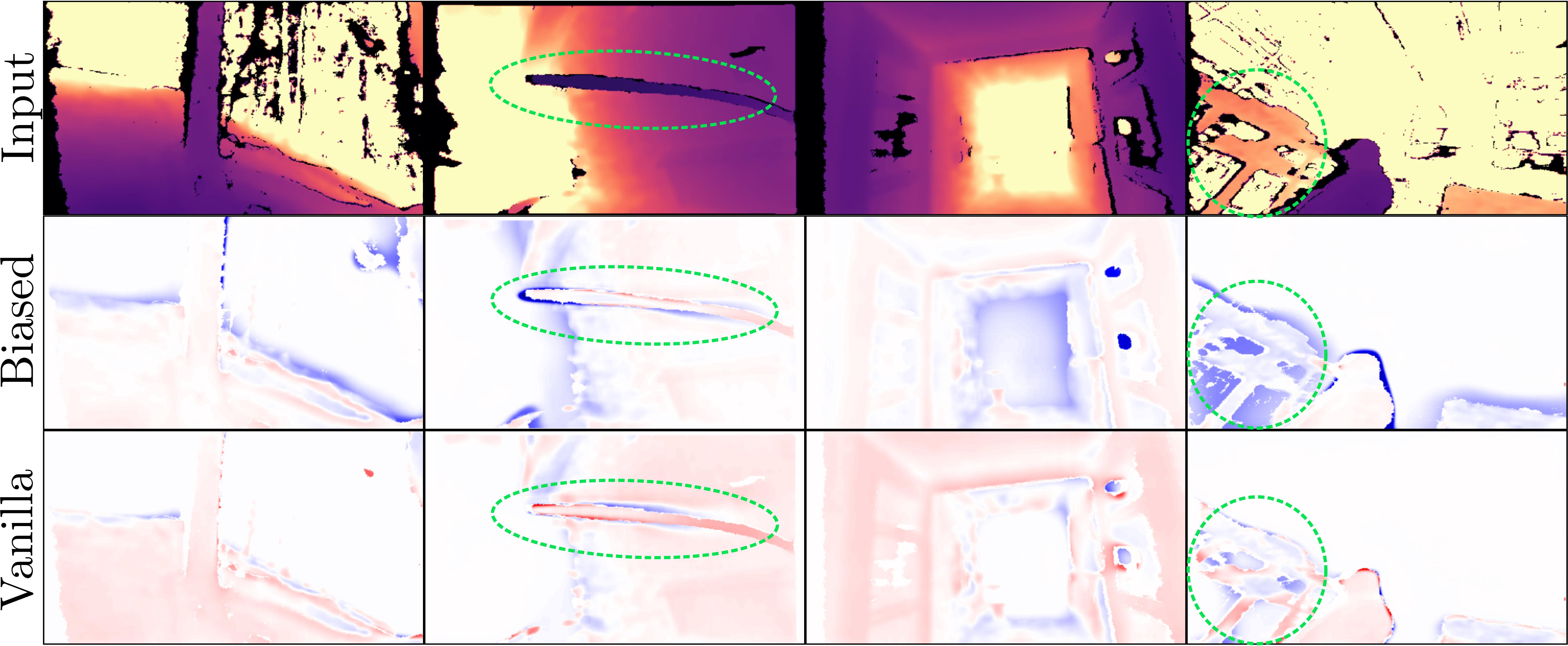}
    \caption{Reconstruction error using the proposed biased VAE and a vanilla VAE.
    The blue pixels correspond to reconstructions ``closer'' than the actual pixel value.
    The green circles highlight instances of thin obstacles whose reconstructions are improved.}
    \label{fig:valid:vae}
\end{figure}

The biased VAE tends to reconstruct pixels closer to the camera.
This is illustrated in selected instances in \fig{fig:valid:vae}.
This results in a conservative approximation of the obstacles, in particular w.r.t. thin or small obstacles.
This (even partial) reconstruction of an obstacle indicates that it is encoded in the latent representation, and therefore can be captured by the SDF network.

To assess generalization, we compute the reconstruction error on depth images from the TartanAir dataset~\cite{Wang20}.
A total of $30,000$ randomly sampled images are evaluated.
These are cropped at $d_\text{max} = \SI{5}{m}$, and images with little to no nearby content (within $d_\text{max}$) are excluded to ensure a representative sample.

The resulting pixel-wise RMSE is \SI{0.225}{m}, which is consistent with the error magnitudes reported in \tab{tab:valid:vae} for our testing depth images.
Specifically, this value is slightly lower as the TartanAir environments typically feature lower spatial frequency than the randomized training scenes.

The average inference time of the VAE encoder, respectively on CPU (Intel Core i7-12800H) and GPU (NVIDIA GeForce RTX 3080 Ti Laptop), is \SI{84.03}{ms} and \SI{1.27}{ms}.
This highlights the importance of GPU acceleration for real-time deployment.

\begin{table}[t]
\centering
    \small
\begin{tabular}{c|c|c|c|c}
    \toprule
          \textbf{Sensor} & \textbf{\makecell{Pixels \\ considered}} & \textbf{FFT} & \textbf{\makecell{Vanilla\\VAE}} & \textbf{\makecell{Biased\\VAE}} \\
    \midrule
        \multirow{2}{*}{{\makecell{Depth \\ camera}}} &
        full image & 0.358 & \textbf{0.182} & 0.313 \\
        & non-background & 0.562 & 0.489 & \textbf{0.342} \\
    \midrule
        \multirow{2}{*}{{LiDAR}} &
        full image & 0.640 & \textbf{0.375} & 0.681 \\
        & non-background & 0.850 & 0.745 & \textbf{0.551} \\
    \bottomrule
\end{tabular}
\normalsize

    \caption{Pixel-wise RMSE [\si{m}] using a FFT, a standard VAE, and the proposed biased VAE,
        on the LiDAR and Depth Camera datasets,
        both on the full image and on the non-background pixels.}
\label{tab:valid:vae}
\end{table}

\subsection{SDF Reconstruction}\label{sec:valid:sdf}

We now evaluate the SDF-approximating MLP.

\subsubsection{Neural Network Size}

We assess how the reconstruction errors vary with the number of neurons.
Network size is indeed a critical parameter, as it directly affects both the solving time of the NMPC,
and the quality of the obstacle avoidance (through the accuracy of the environment encoding).

\begin{table}[t]
\centering
    \small
\setlength{\tabcolsep}{3pt}
\begin{tabular}{c|cccccc}
    \toprule
        \textbf{Abbrev.}
        & \rotatebox[origin=c]{65}{SDF${}_{64}$}
        & \rotatebox[origin=c]{65}{SDF${}_{128}$}
        & \rotatebox[origin=c]{65}{\textbf{SDF}$\mathbf{{}_{256-64}}$}
        & \rotatebox[origin=c]{65}{SDF${}_{256}$}
        & \rotatebox[origin=c]{65}{SDF${}_{512-256}$}
        & \rotatebox[origin=c]{65}{\textbf{SDF}$\mathbf{{}_{512}}$} \\
    \midrule
        \textbf{\makecell{Layer\\sizes}}
        \footnotesize
        & $\smat{64\\64\\64\\64}$
        & $\smat{128\\128\\128\\128}$
        & $\smat{256\\256\\128\\64}$
        & $\smat{256\\256\\256\\256}$
        & $\smat{512\\512\\256\\256}$
        & $\smat{512\\512\\512\\512}$
        \small \\
    \midrule
    \textbf{\makecell{\# of\\params}}
        & $41217$
        & $115201$
        & $246401$
        & $361473$
        & $869889$
        & $1247233$ \\
    \midrule
        \textbf{\makecell{Query\\time}}
        & $0.53$
        & $0.66$
        & $0.69$
        & $0.73$
        & $0.89$
        & $1.06$ \\
    \bottomrule
\end{tabular}
\normalsize
\setlength{\tabcolsep}{6pt}

    \caption{Details of the $6$ evaluated network sizes, with their naming abbreviations.
    The average query time [\si{ms}] for the SDF value and the corresponding gradient on a single query point is evaluated on an Intel Core i7-12800H CPU.
    The two networks highlighted in bold are further evaluated for closed-loop performances in \sect{sec:valid:ablation}.}
\label{tab:valid:sdf_sizes}
\end{table}

\begin{figure}[t]
\centering
    \includegraphics[width=\columnwidth]{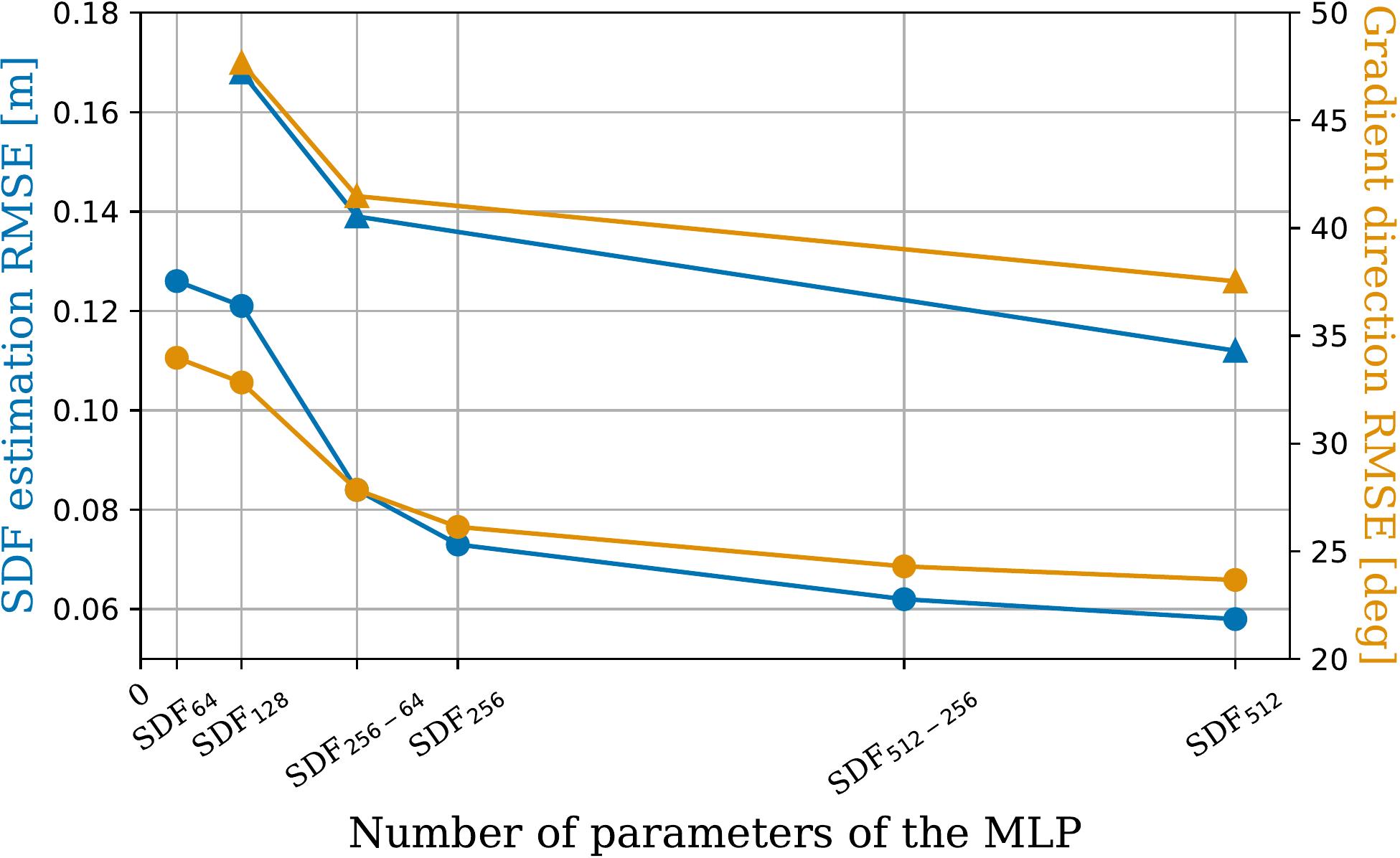}
    \caption{RMSE of the SDF estimation (blue) and of its gradient orientation (orange),
        for the depth (circles) and LiDAR (triangles) images,
        as a function of the neural network size.}
    \label{fig:valid:sdf_metrics}
\end{figure}

Specifically, we evaluate $6$ sets of layer sizes,
and report the corresponding SDF reconstruction errors (and gradient direction errors) in \fig{fig:valid:sdf_metrics}.
The evaluated network sizes, detailed in \tab{tab:valid:sdf_sizes}, are named after the following convention:
SDF${}_{N}$ denotes a network where all four layers have $N$ neurons,
and SDF${}_{N-M}$ denotes a network with decreasing layer sizes from $N$ in the first layer to $M$ in the last.
We chose as the smallest architecture SDF${}_{64}$ the one used in~\cite{Jacquet24},
which has been shown successful in encoding an occupancy map, \ie, a different (simpler) spatial representation.
The network size SDF${}_{256}$ is similar to the one used in~\cite{Ortiz22} for online learning of a weight-encoded SDF.
Metrics are computed on a 3D grid with a resolution of \SI{10}{cm} within the sensor frustum.
The evaluation is performed solely on simulated data, as the ground truth cannot be obtained for images that contain invalid pixels.
It can be observed that the RMSE is higher in the LiDAR case,
which is consistent with the VAE results in \tab{tab:valid:vae}.

\begin{figure}[t]
\centering
    \includegraphics[width=\columnwidth]{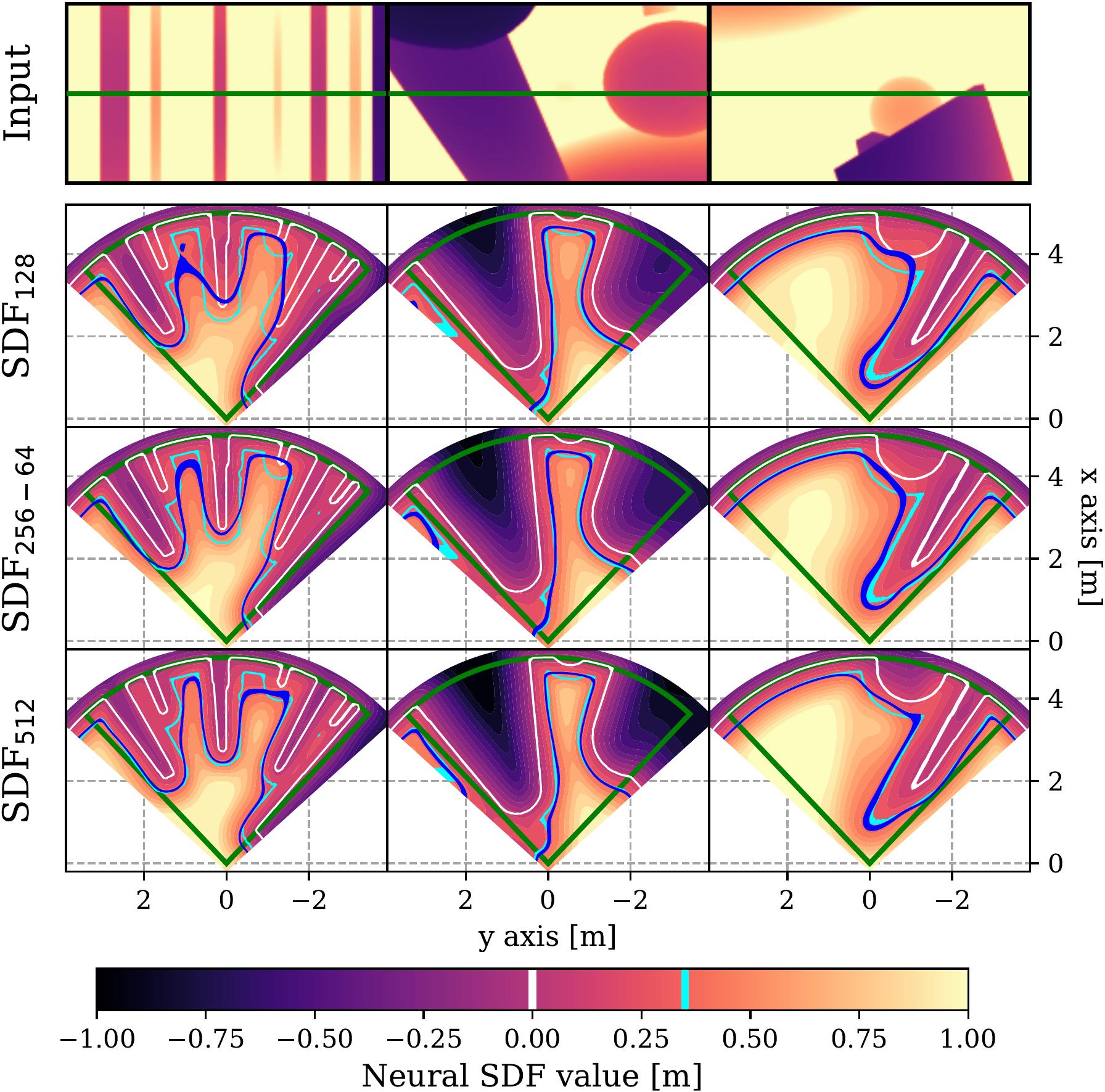}
    \caption{$2$D slice of the estimated SDF at $z=0$ (shown as the green line).
        The green cones depict the FoV.
        The color gradient shows the neural SDF,
        and the white curve marks the visible obstacle surface.
        The blue and cyan curves are the neural $(r+\epsilon)$-level sets, used as constraint, and its ground truth, respectively.}
    \label{fig:valid:sdf}
\end{figure}

In order to provide visual insights on the neural $3$D field, some instances of $2$D SDF reconstruction are depicted in \fig{fig:valid:sdf} for the slices $z_B=0$.
It illustrates the convergence of the network-predicted SDF $(r + \epsilon)$-level set (blue) toward the ground truth (cyan) as network size increases. Notably, the SDF${}_{128}$ architecture exhibits significant penetration into obstacle, rendering it unsuitable for navigation tasks.
Indeed, \fig{fig:valid:sdf_metrics} shows a strong error decrease between SDF${}_{128}$ and SDF${}_{256-64}$.
We select SDF${}_{256-64}$ as the default architecture for the remainder of the paper, as it provides a practical trade-off between computational efficiency and reconstruction accuracy.
For comparison, we also evaluate SDF${}_{512}$ to explore how further accuracy improvements translate to closed-loop navigation performance.

\subsubsection{Comparison with Other SDF Approximations}

To contextualize the performance of our method relative to existing SDF encoding approaches, we compare computation times in \tab{tab:valid:cpt}.
Specifically, we benchmark against a GP-based method~\cite{Wu23},
implemented using the GPU-accelerated GPyTorch library~\cite{Gardner18}.
As an additional baseline, we include a k-d tree implementation.
Although it does not provide an analytical SDF and thus cannot be used in our framework, it serves as a useful point of reference for raw query performance.

\begin{table}[t]
\centering
    \small
\begin{tabular}{c|cc|cc|cc}
    \toprule
        \multirow{3}{*}{\textbf{\makecell{Input \\ size}}} &
        \multicolumn{2}{c|}{\textbf{k-d tree}} &
        \multicolumn{2}{c|}{\textbf{GP}} &
        \multicolumn{2}{c}{\textbf{neural SDF}} \\
        &
        \footnotesize{constr.} & \footnotesize{query} &
        \footnotesize{constr.} & \footnotesize{query} &
        \footnotesize{VAE} & \footnotesize{query} \\
        &
        \footnotesize{(CPU)} & \footnotesize{(CPU)} &
        \footnotesize{(GPU)} & \footnotesize{(CPU)} &
        \footnotesize{(GPU)} & \footnotesize{(CPU)} \\
    \midrule
        129600 &
        41.0 & 0.5 &
        - & - &
        1.3 & 0.7 \\
        14400 &
        4.1 & 0.2 &
        587.3 & 2.4 &
        - & - \\
        1296 &
        0.3 & 0.1 &
        38.7 & 2.3 &
        - & - \\
        190 &
        0.1 & 0.1 &
        5.5 & 1.9 &
        - &  - \\
    \bottomrule
\end{tabular}
\normalsize

    \caption{Computation times [\unit{ms}] for the construction and query (of the value and gradient) for different SDF approximation methods applied to depth images, with varying input dimensions.}
\label{tab:valid:cpt}
\end{table}

While these methods compute a different SDF representation, preventing a direct one-to-one comparison, we include the reconstruction errors reported in~\cite{Wu23} and~\cite{Ortiz22} for reference alongside the values reported in~\fig{fig:valid:sdf}.
The former reports an instance of scene RMSE of \SI{7.7}{cm},
and the latter reports errors between \num{3} and \SI{7}{cm} across different scenes.
These values are comparable in scale to the RMSE obtained with our neural SDF (albeit generally lower).
The gradient error reported in~\cite{Ortiz22} is also comparable, lying between 25\degree and 30\degree.

\subsubsection{Generalizability}

We further evaluate the SDF reconstruction error on depth images from the TartanAir dataset~\cite{Wang20}.
Using the same $30,000$ images as in \sect{sec:valid:vae},
we compute the reconstruction error with the SDF${}_{256\text{-}64}$ network, on a 3D grid with a resolution of \SI{10}{cm}.
The recorded RMSE is \SI{9.1}{cm} ($+8\%$) for the SDF estimation, and the gradient direction errors is $31$\degree ($+10\%$), both remaining within the same range as those reported in~\fig{fig:valid:sdf_metrics} and in related works.

\subsection{Ablation Study}\label{sec:valid:ablation}

\begin{table*}[t]
\centering
    \newcounter{linecounter}
\setcounter{linecounter}{1}
\newcommand{\linecount}[0]{
    \ifnum\value{linecounter}<10 0\fi\arabic{linecounter})~
    \stepcounter{linecounter}
}

\small
\begin{tabular}{|c|l|c|c|c|c|c|c|}
    \hline
         &
        \makecell{\centering\textbf{\begin{tabular}[c]{@{}c@{}}Altered\\ parameters\end{tabular}}} &
        \textbf{\begin{tabular}[c]{@{}c@{}}Success\\ Rate\end{tabular}} &
        \textbf{\begin{tabular}[c]{@{}c@{}}Timeout\\ Rate\end{tabular}} &
        \textbf{\begin{tabular}[c]{@{}c@{}}Failure\\ Rate\end{tabular}} &
        \textbf{\begin{tabular}[c]{@{}c@{}}Avg.\\ Velocity\end{tabular}} &
        \textbf{\begin{tabular}[c]{@{}c@{}}Avg. min.\\ neural SDF\end{tabular}} &
        \textbf{\begin{tabular}[c]{@{}c@{}}Avg. min.\\ SDF\end{tabular}} \\
    \hline
    \hline
        \multirow{20}{*}{\rotatebox[origin=c]{90}{\textbf{\makecell{Pillars}}}} &
        \textit{Baseline} &
        1.0 &
        0 &
        0 &
        1.75 &
        0.339 &
        0.446 
    \\ \cline{2-8}
         &
        \linecount sensor frequency $2$\unit{Hz} &
        1.0 &
        0 &
        0 &
        1.74 &
        0.359 &
        0.482 
    \\ \cline{2-8}
         &
        \linecount higher state noise ($\sigma=0.05$) &
        0.985 &
        0 &
        0.015 &
        1.70 &
        0.378 &
        0.473 
    \\ \cline{2-8}
         &
        \linecount higher state noise ($\sigma=0.07$) &
        0.96 &
        0 &
        0.04 &
        1.59 &
        0.508 &
        0.441 
    \\ \cline{2-8}
         &
        \linecount $v_\text{ref} = 3$\unit{m/s} &
        0.93 &
        0 &
        0.07 &
        2.34 &
        0.368 &
        0.464 
    \\ \cline{2-8}
         &
        \linecount $v_\text{ref} = 3$\unit{m/s} , SDF${}_{512}$ &
        0.97 &
        0 &
        0.03 &
        2.34 &
        0.368 &
        0.464 
    \\ \cline{2-8}
         &
        \linecount $\alpha_H = 65$\degree &
        0.995 &
        0 &
        0.005 &
        1.63 &
        0.236 &
        0.376 
    \\ \cline{2-8}
         &
        \linecount $\alpha_H = 65$\degree , $v_\text{ref} = 3$\unit{m/s} &
        0.95 &
        0 &
        0.05 &
        2.1 &
        0.288 &
        0.413 
    \\ \cline{2-8}
         &
        \linecount $\alpha_H = 65$\degree , $v_\text{ref} = 3$\unit{m/s} , SDF${}_{512}$ &
        0.99 &
        0 &
        0.01 &
        2.18 &
        0.333 &
        0.392 
    \\ \cline{2-8}
         &
        \linecount $\alpha_H = 180$\degree &
        0.99 &
        0 &
        0.01 &
        1.48 &
        0.303 &
        0.350 
    \\ \cline{2-8}
         &
        \linecount $\alpha_H = 180$\degree , $v_\text{ref} = 3$\unit{m/s} &
        0.945 &
        0 &
        0.055 &
        2.18 &
        0.321 &
        0.361 
    \\ \cline{2-8}
         &
        \linecount $\alpha_H = 180$\degree , $v_\text{ref} = 3$\unit{m/s} , SDF${}_{512}$ &
        0.99 &
        0 &
        0.01 &
        2.05 &
        0.297 &
        0.354 
    \\ \cline{2-8}
         &
        \linecount $d_\text{min}=0.80$\unit{m} &
        0.92 &
        0 &
        0.08 &
        1.53 &
        0.304 &
        0.419 
    \\ \cline{2-8}
         &
        \linecount $d_\text{min}=0.75$\unit{m} &
        0.77 &
        0.03 &
        0.2 &
        1.41 &
        0.304 &
        0.419 
    \\ \cline{2-8}
         &
        \linecount $d_\text{min}=0.80$\unit{m} , $\alpha_H = 65$\degree &
        0.97 &
        0 &
        0.03 &
        1.35 &
        0.236 &
        0.376 
    \\ \cline{2-8}
         &
        \linecount $d_\text{min}=0.75$\unit{m} , SDF${}_{512}$ &
        0.95 &
        0 &
        0.05 &
        1.64 &
        0.352 &
        0.406 
    \\ \cline{2-8}
         &
        \linecount $d_\text{min}=0.75$\unit{m} , $\alpha_H = 65$\degree &
        0.88 &
        0.05 &
        0.07 &
        1.23 &
        0.220 &
        0.359 
    \\ \cline{2-8}
         &
        \linecount $d_\text{min}=0.75$\unit{m} , $\alpha_H = 65$\degree , SDF${}_{512}$ &
        1.0 &
        0 &
        0 &
        1.47 &
        0.272 &
        0.343 
    \\ \cline{2-8}
         &
        \linecount $d_\text{min}=0.75$\unit{m} , $\alpha_H = 180$\degree &
        0.215 &
        0.78 &
        0.05 &
        0.83 &
        0.258 &
        0.305 
    \\ \cline{2-8}
         &
        \linecount $d_\text{min}=0.75$\unit{m} , $\alpha_H = 180$\degree , SDF${}_{512}$ &
        1.0 &
        0 &
        0 &
        1.42 &
        0.277 &
        0.295 
    \\ \cline{2-8}
         &
        \linecount $d_\text{min}=0.75$\unit{m} , $v_\text{ref} = 1$\unit{m/s} &
        0.94 &
        0 &
        0.06 &
        1.09 &
        0.421 &
        0.374 
\\
\hline
\hline

        \multirow{14}{*}{\rotatebox[origin=c]{90}{\textbf{\makecell{Random}}}} &
        \textit{Baseline} &
        0.88 &
        0.065 &
        0.055 &
        1.72 &
        0.288 &
        0.345 
    \\ \cline{2-8}
         &
        \linecount $v_\text{ref} = 3$\unit{m/s} &
        0.88 &
        0.02 &
        0.1 &
        2.43 &
        0.419 &
        0.465 
    \\ \cline{2-8}
         &
        \linecount SDF${}_{512}$ &
        0.905 &
        0.05 &
        0.045 &
        1.76 &
        0.453 &
        0.442 
    \\ \cline{2-8}
         &
        \linecount $\alpha_H = 65$\degree &
        0.925 &
        0.03 &
        0.045 &
        1.71 &
        0.298 &
        0.352 
    \\ \cline{2-8}
         &
        \linecount $\alpha_H = 65$\degree , $v_\text{ref} = 1$\unit{m/s} &
        0.885 &
        0.095 &
        0.02 &
        1.13 &
        0.207 &
        0.252 
    \\ \cline{2-8}
         &
        \linecount $\alpha_H = 65$\degree , SDF${}_{512}$ &
        0.925 &
        0.035 &
        0.04 &
        1.67 &
        0.301 &
        0.320 
    \\ \cline{2-8}
         &
        \linecount $\alpha_H = 180$\degree &
        0.95 &
        0.05 &
        0 &
        1.70 &
        0.241 &
        0.310 
    \\ \cline{2-8}
         &
        \linecount $\alpha_H = 180$\degree , SDF${}_{512}$ &
        0.97 &
        0.025 &
        0.05 &
        1.65 &
        0.256 &
        0.402 
    \\ \cline{2-8}
         &
        \linecount $+20$ obstacles &
        0.87 &
        0.05 &
        0.08 &
        1.70 &
        0.454 &
        0.450 
    \\ \cline{2-8}
         &
        \linecount $+20$ thin obstacles &
        0.82 &
        0.05 &
        0.13 &
        1.68 &
        0.448 &
        0.421 
    \\ \cline{2-8}
         &
        \linecount $+20$ thin obstacles , SDF${}_{512}$&
        0.88 &
        0.03 &
        0.09 &
        1.72 &
        0.435 &
        0.420 
    \\ \cline{2-8}
         &
        \linecount $+20$ thin obstacles , $\alpha_H = 65$\degree &
        0.88 &
        0.03 &
        0.09 &
        1.69 &
        0.258 &
        0.296 
    \\ \cline{2-8}
         &
        \linecount $+20$ thin obstacles , $\alpha_H = 180$\degree &
        0.9 &
        0.03 &
        0.07 &
        1.56 &
        0.240 &
        0.355 
    \\ \cline{2-8}
         &
        \linecount $+20$ thin obstacles , $\alpha_H = 180$\degree , SDF${}_{512}$&
        0.92 &
        0.03 &
        0.05 &
        1.61 &
        0.234 &
        0.368 
    \\ \hline
\end{tabular}
\normalsize

    \caption{Evaluation metrics in the two classes of environments in various simulation settings.
        Each line corresponds to metrics gathered over 200 randomized rollouts for a single set of parameters,
        enumerated and described in the first column.}
    \label{tab:valid:ablation}
    \vspace{-2ex}
\end{table*}

In this section, we evaluate the sensitivity of performances w.r.t. the various hyperparameters.
Although NMPC is typically paired with a low-level feedback controller for added robustness against model mismatches, we isolate our analysis by modifying the NMPC to output body wrench commands directly.
We perform the simulations in the Aerial Gym simulator~\cite{Kulkarni25},
which supports customizable physics stepping frequencies, randomized obstacle generation, and parallel rollouts.
Control is executed at \SI{50}{Hz}, while the physics engine runs at \SI{500}{Hz}.

The simulated AR weights \SI{1.25}{kg} and has a radius $r=\SI{25}{cm}$.
We fix the safety margin $\epsilon=\SI{10}{cm}$.
Each rollout samples a new obstacle configuration within a $10 \times 10 \times \SI{5}{m}$ environment, as well as the start and goal locations on opposite sides.
Rollouts terminate colliding with an obstacle, or timing out after \SI{30}{s}, if the goal is not reached, indicating that the robot is stuck in a dead-end.
This behavior is expected in highly cluttered scenarios, as the planner only has access to local information.

Performance metrics include success, failure, and timeout rates,
as well as other metrics indicating the navigation performance:
average speed,
average minimum neural SDF evaluation during the rollout,
the true SDF values based on range observations.
The robot consistently reaches its target speed for a significant portion of time; specifically, we verify that the average $90$\ith percentile speed remains above $0.95 v_\text{ref}$.

\begin{figure}[t]
\centering
\newlength\figheight
\setlength{\figheight}{2.9cm}
    \includegraphics[height=\figheight]{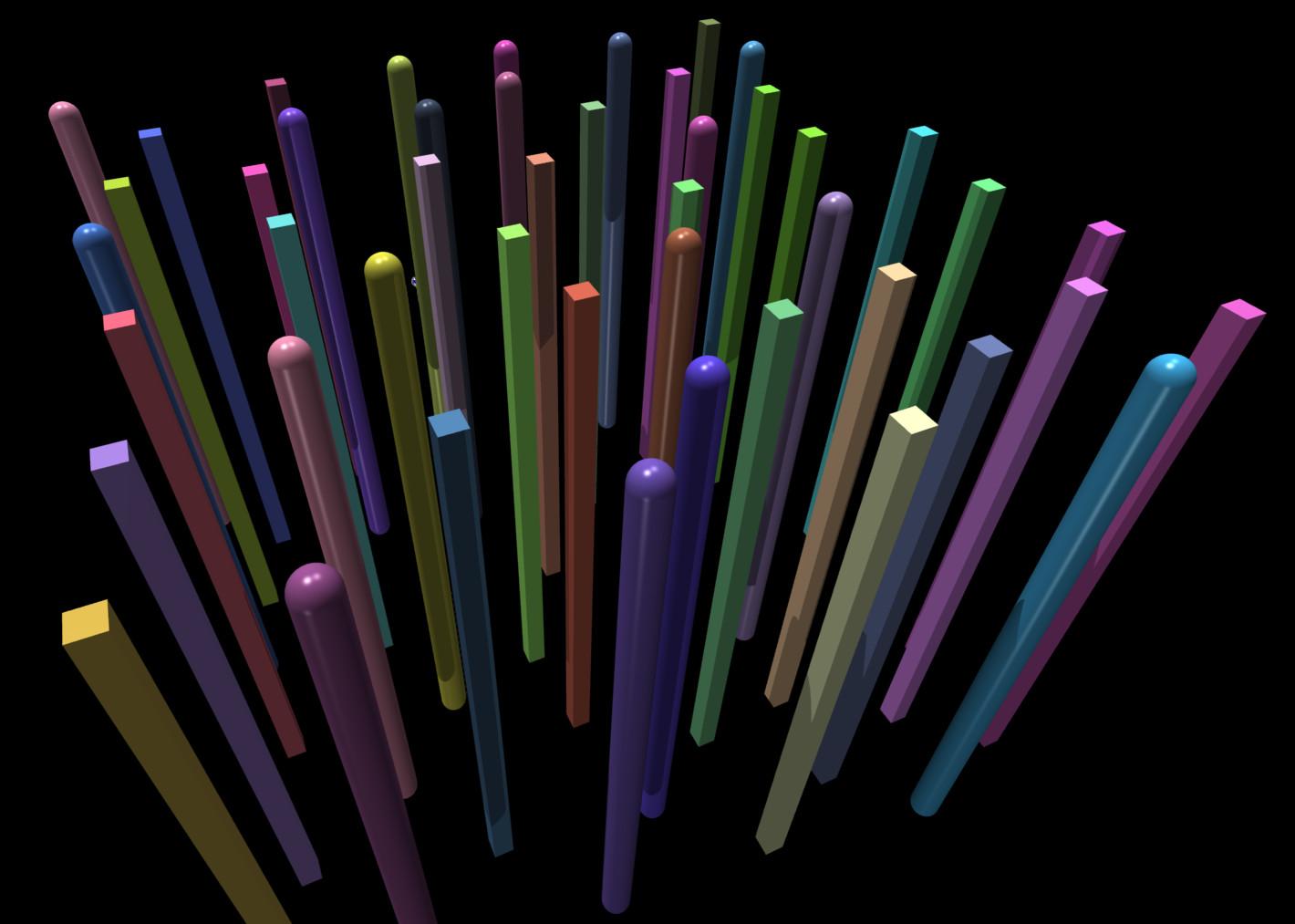}
    \hfill
    \includegraphics[height=\figheight]{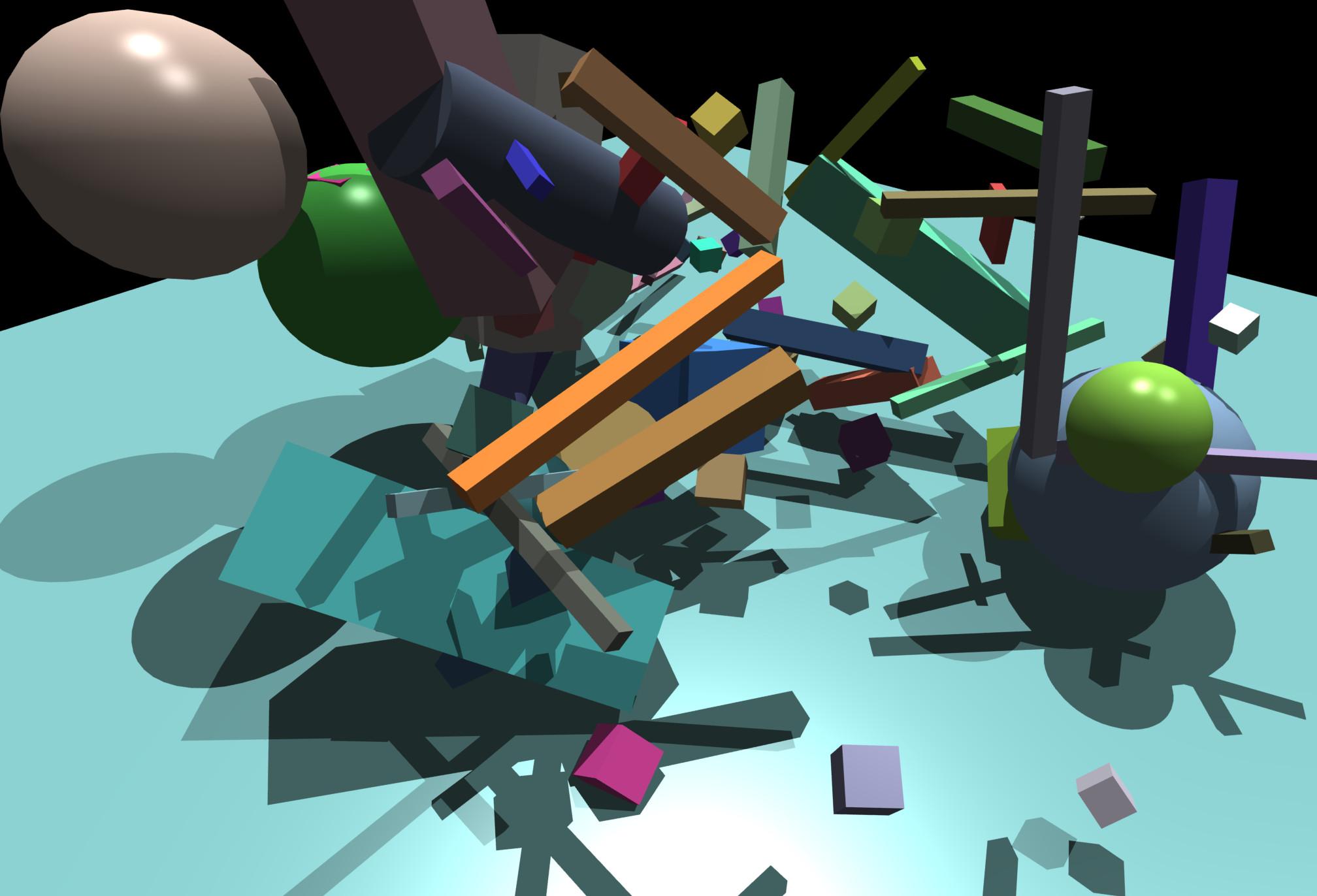}
    \caption{Instances of the two classes of environments used in the ablation study.}
    \label{fig:valid:env_examples}
    \vspace{-2ex}
\end{figure}

We consider two classes of environments, pictured in \fig{fig:valid:env_examples}.
First, the obstacles are vertical pillars of circular or square sections, with diameters (or diagonals) ranging from $0.2$ to $0.4$\si{m}.
These are sampled using Poisson discs, with a tunable minimum inter-obstacle distance $d_\text{min}$.
This setup creates an effectively 2D navigation problem, enabling easier qualitative interpretation.
Second, we use $3$D randomly sampled cuboids, spheres, pillars, and rods, with smallest dimensions as low as \SI{0.2}{m}.
We also assess the impact of thinner obstacles (with smallest dimensions $<\SI{0.05}{m}$), such as small cuboids and rods.

The baseline conditions are defined after the Intel D455 depth camera FoV (\ie, $\alpha_H=45$\degree) at \SI{25}{Hz},
the SDF${}_{256-64}$ network and $v_\text{ref} = \SI{2}{m/s}$.
Gaussian noise is applied on the state feedback provided to the NMPC and on the Gaussian noise is applied to both state feedback and sensor observations (with stds $\sigma = 0.03$ and $\sigma = 0.05$, respectively),
and control wrench perturbations are also introduced.
Rollout visualizations for both environment classes are shown in Extension 1.
Results are reported in \tab{tab:valid:ablation} and discussed below.

\paragraph{Pillar Environments}
First, we observe (lines 1-3) that the framework exhibits resilience to reduced sensor frequency and moderate state noise—maintaining failure rates under 4\% even with noisy feedback.
The parameters that critically affect the success metrics are the reference velocity, the obstacle density, and the SDF network size.
At higher velocities (lines 4-5, 7-8, and 10-11), reduced reaction times increase sensitivity to SDF approximation errors.
This is mitigated by using the SDF${}_{512}$ network, which enables sharper reconstructions (see \fig{fig:valid:sdf}).
In the simple baseline environment (lines 1-11),
the FoV only marginally impacts the performance.
However, increasing the obstacle density (lines 12-20) causes a drop in success (down to $20\%$ failure rate, line 13).
This is attributed to the slackened constraint~\eqref{eq:mpc:nlp:hfov_constr} which allows slight lateral drift and results in collisions in denser layouts.
A larger FoV mitigates this issue.
It can be noted that the baseline SDF network struggles to reconstruct properly the narrow gaps in dense environments,
especially for large FoV (line 18).
This leads the neural constraint to prevents motion entirely, causing timeouts despite no actual dead-ends being present (since we ensure $d_\text{min} > 2(r + \epsilon) = \SI{0.7}{m}$).

\begin{figure*}[t]
    \includegraphics[width=\linewidth]{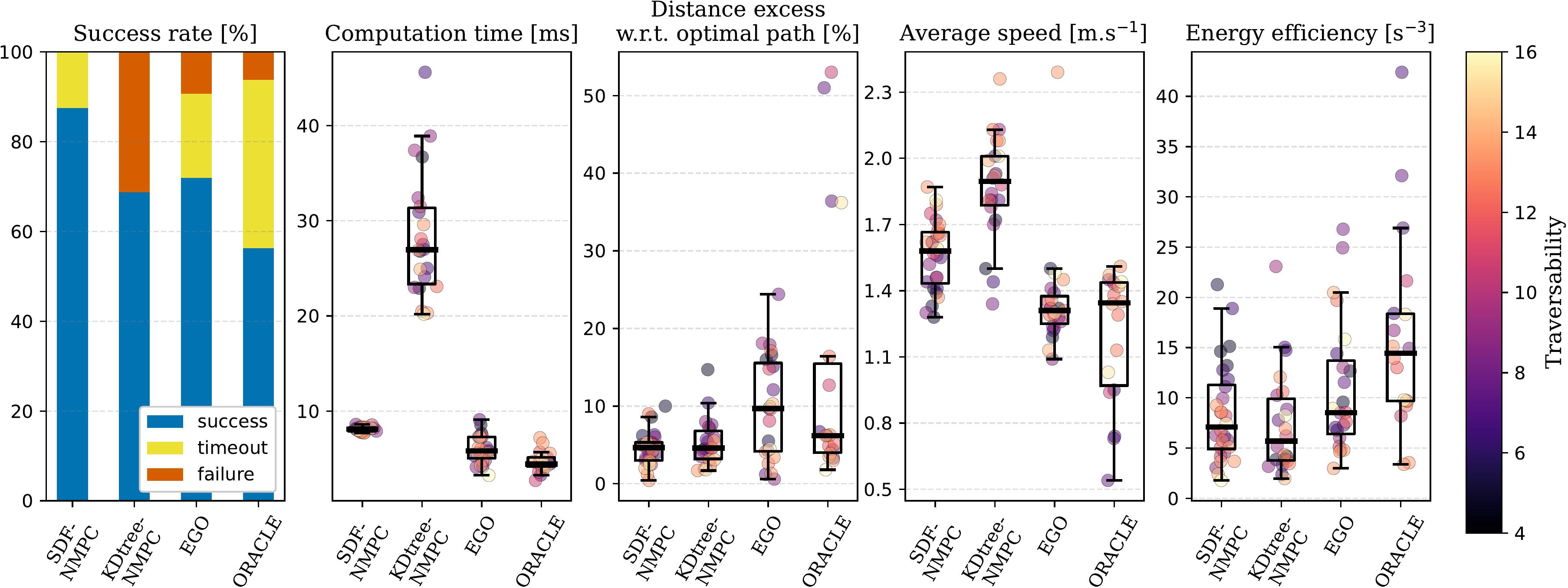}
    \caption{Comparative metrics for the 4 evaluated methods.
        The boxplots are overlayed with the individual samples for each of the successful rollouts,
        colored by traversability, as defined in~\cite{Yu23}.}
    \label{fig:valid:comp}
\end{figure*}

\paragraph{3D Random Clutter}
Simulations in 3D environments show similar trends, though baseline failure rates are non-zero ($5.5$\%), reflecting the known difficulty of mapless methods in cluttered settings with limited FoV.
Slack FoV constraints also contribute to lateral collisions in dense areas.
In such unstructured environments, increasing the FoV dramatically improves the success rate (lines 23-27), approaching $100\%$ success rate in the $360$\degree FoV case (lines 26-27).
The framework also scales favorably with increasing obstacle density (lines 28),
but performance degrades in the presence of thin obstacles (lines 29-33) (from $8\%$ to $13\%$ failure rate).
Thin obstacles are challenging to encode and reconstruct, especially with the smaller SDF${}{256-64}$ network.
The larger SDF${}{512}$ improves results (line 33), reducing the failure rate to 5\%.

Across all simulations, the final columns of \tab{tab:valid:ablation} indicate that the SDF network is generally conservative in its distance estimates.
This suggests that most SDF constraint violations stem from discontinuities in the observations and poor estimations in the SDF network.

The average NMPC solving time on an Intel Core i7-12800H CPU is \SI{8.51}{ms} using SDF${}_{256-64}$, and \SI{15.18}{ms} using SDF${}_{512}$.

\subsection{Comparison with Existing Methods}\label{sec:valid:comparison}

In this section, the proposed SDF-NMPC method is evaluated against three state-of-the-art collision avoidance approaches based on local depth observations, both neural and non-neural:
\begin{itemize}
    \item the spline-based EGO-Planner from~\cite{Zhou21a} (and more specifically, the improved implementation EGO-Swarm \cite{Zhou21b, Zhou22}),
    \item the neural, motion-primitive-based collision predictor ORACLE from~\cite{Nguyen24},
    \item and the k-d tree-based NMPC from~\cite{Zhang25}.
\end{itemize}

A key distinction of our method is its formal consideration of safety as an explicit constraint in the optimization, rather than the heuristic collision score in ORACLE or the optimization co-objective in EGO and KDtree-NMPC.

The comparison is conducted using the Flightmare simulator~\cite{Song21} and benchmarking metrics from in~\cite{Yu23}:
\begin{enumerate}
    \item the path optimality, \ie the ratio of excess traveled distance relative to an A* baseline,
    \item the average speed, normalized by the optimal path length,
    \item and the energy efficiency (\ie, the integral of the jerk).
\end{enumerate}

All methods are evaluated in the same 32 randomly sampled environments, both indoor and outdoor~\cite{Yu23}.
The obstacles are sampled through Poisson disc distributions with radii chosen to cover a large range of traversability ($\approx4$ to $16$).
The target navigation speed is \SI{2}{m/s}.

Baseline parameters are drawn from publicly available implementations, and fine-tuned for performance in the evaluation environments.
Simulations are performed on a workstation equipped with an NVIDIA GeForce RTX 3090 GPU and an AMD Ryzen Threadripper 3970X CPU.
A visualization of the simulations is included in Extension 1, and results are reported in \fig{fig:valid:comp}.

The proposed NMPC method (using SDF${}_{256-64}$) outperforms the other methods in terms of success rate, achieving zero collisions.
While being more computationally demanding than EGO and ORACLE, it maintains a control frequency above \SI{100}{Hz} and is more efficient than the KDtree-NMPC.
Notably, its computation time is more consistent around the average, whereas the baselines show increased computational load as traversability decreases.

In terms of path optimality, both our NMPC and KDtree-NMPC outperform EGO and ORACLE, benefiting from objectives of minimizing deviation from a straight path.
Although KDtree-NMPC achieves higher speeds, our method still surpasses EGO and ORACLE in this regard, demonstrating strong tracking performance without compromising safety.

Finally, energy efficiency is comparable across most methods, except for ORACLE, whose switching acceleration and yawing rate outputs do not account for dynamic feasibility and smoothness.

\subsection{Resilience to Drifting Odometry}\label{sec:valid:drift_odom}

Because the collision avoidance is formulated in the local frame, our framework is resilient to drifting odometry, which would hinder the construction of a reliable map.
In this section, we present an ablation study evaluating the method’s resilience to increasing odometry drift and decreasing sensor frequency.
The latter is relevant since, without new measurements, the method relies on the forward-propagated last received measurement; thus, a lower frequency implies a higher sensitivity to drift.

In this study, we perform simulations in Gazebo, and deteriorate the state feedback provided to the NMPC.
Specifically, velocity and yaw rate are perturbed using Gaussian noise and a random-walk bias.
Position and heading are then obtained through integration.
The tilt angles are not altered, as those are non-drifting quantities directly observable using an IMU.

The parameters of these noises are empirically selected to achieve the desired Relative Position Error (RPE)~\cite{Geiger12} (computed for a delta of \SI{10}{m}).
This metric computes the position RMSE per delta of traveled distance.
It offers a more meaningful measure of drift than absolute position error.
For reference, the Visual-Inertial Odometry (VIO) methods ROVIO~\cite{Bloesch17} and VINS-Mono~\cite{Qin18} both report RPE of $1$ to $2\%$ for the same delta.

\begin{figure}[t]
\centering
    \includegraphics[width=\columnwidth]{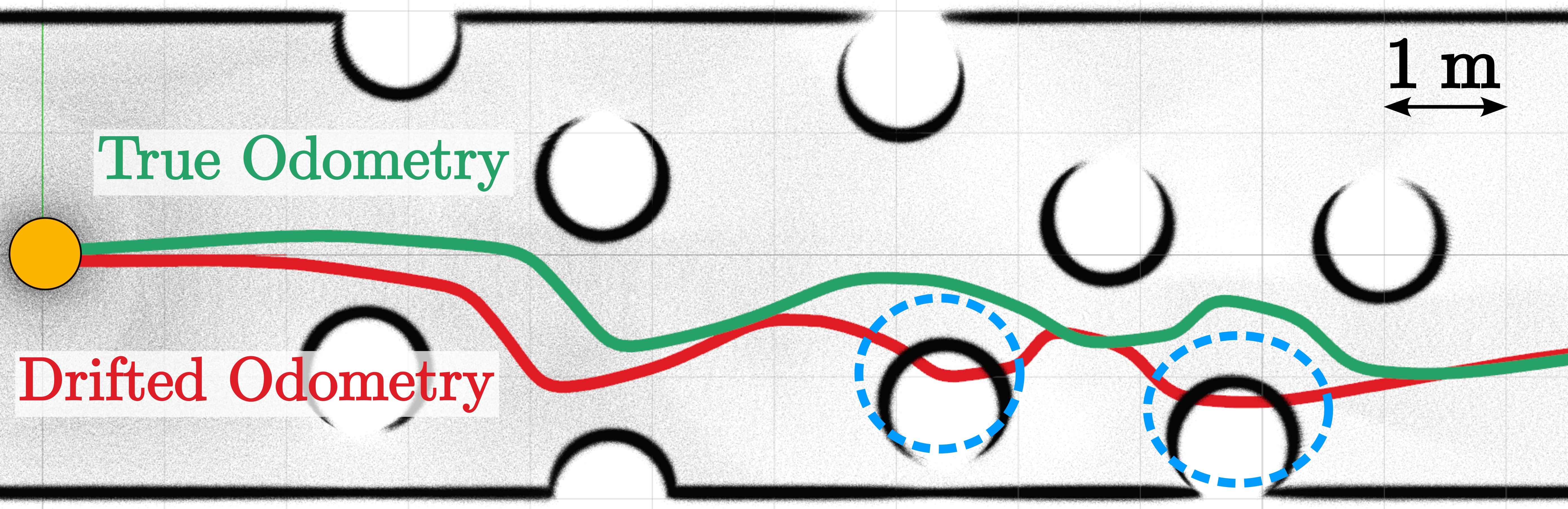}
    \caption{Top-view of a $2$D trajectory where the NMPC receives drifting odometry (red).
    Despite the drift,  the real system (green trajectory) avoids collisions, as the controller operates in a local observation frame independent of map-based positioning.
        The orange circle depicts the robot size.}
\label{fig:valid:drift_sim_env}%
\end{figure}

We perform these simulations in random, cluttered environments where spheres of \SI{1}{m} radius are sampled using 3D Poisson Discs with a radius of \SI{1.5}{m}.
The environment is \SI{50}{m} long, enclosed by walls, and the robot must reach a waypoint on the opposite end (with a reference speed of \SI{2}{m/s}).
This setup is intentionally challenging due to high clutter and a constrained FoV, making it a rigorous test of the method’s resilience to drift.
An illustrative, $2$D example of such an environment is provided in \fig{fig:valid:drift_sim_env}.
We evaluate the success rate of the NMPC by performing rollouts in each scenario.

\begin{table}[t]
\centering
    \small
\begin{tabular}{cr|ccccc}
    \toprule
        \multicolumn{2}{c|}{} &
        \multicolumn{5}{c}{\textbf{Sensor Frequency}} \\

        \multicolumn{2}{c|}{} &
        \SI{100}{Hz} &
        \SI{30}{Hz} &
        \SI{10}{Hz} &
        \SI{1}{Hz} &
        \SI{0.5}{Hz} \\
    \midrule
        \multirow{4}{*}{\rotatebox[origin=c]{90}{\textbf{RPE}}} &
        $0\%$ &
        10 &
        10 &
        10 &
        10 &
        7 \\

        &
        $\sim5\%$ &
        10 &
        9 &
        8 &
        7 &
        1 \\

        &
        $\sim10\%$ &
        10 &
        8 &
        6 &
        1 &
        0 \\

        &
        $\sim15\%$ &
        7 &
        3 &
        2 &
        0 &
        0 \\
    \bottomrule
\end{tabular}
\normalsize

    \caption{Number of successes out of $10$ rollouts with increasing RPE and decreasing sensor frequency. The reference velocity is \SI{2}{m/s}.}
\label{tab:valid:drifting_odom}%
\end{table}

\tab{tab:valid:drifting_odom} reports the number of successes out of ten ten different environments in each RPE~/~sensor frequency setting.
The first column and first row correspond to the limit cases (perfect odometry and \SI{100}{Hz} sensor frequency -- \ie, the control frequency).
We remark that the success rate is not $10$ out of $10$ in the two extreme cases (top right and bottom left cells).
This is due to a)~the severely degraded velocity estimate in high RPE scenarios that leads to poor predictive performance (the NMPC formulation is not robust);
and b)~the uncertainties in the SDF in the very cluttered environments lead to collisions when the sensor rate becomes too low.

These results highlight the local aspect of the method, which achieves collision avoidance while relying on short-term, local consistency of the state estimate in the weakest sense without explicitly considering robust formulations (\ie, allows forward propagation in between two successive range measurements).

\subsection{Real World Experiments}\label{sec:valid:xp}

\subsubsection{System Setup}\label{sec:valid:xp:setup}

\begin{figure}[t]
\centering
    \includegraphics[width=\columnwidth]{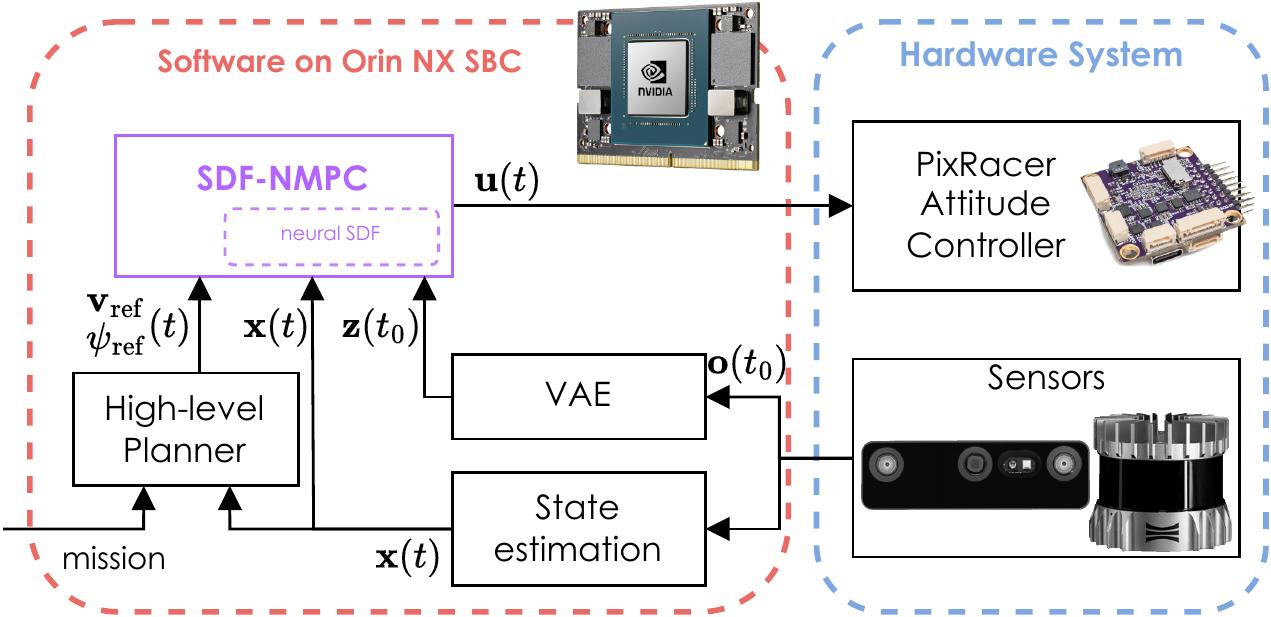}
    \caption{Block diagram on the SDF-NMPC framework. The contributed controller, which integrates the neural SDF network as a constraint, is highlighted in purple.}
    \label{fig:valid:block_diag}
\end{figure}

\begin{figure}[t]
\centering
    \includegraphics[width=\columnwidth]{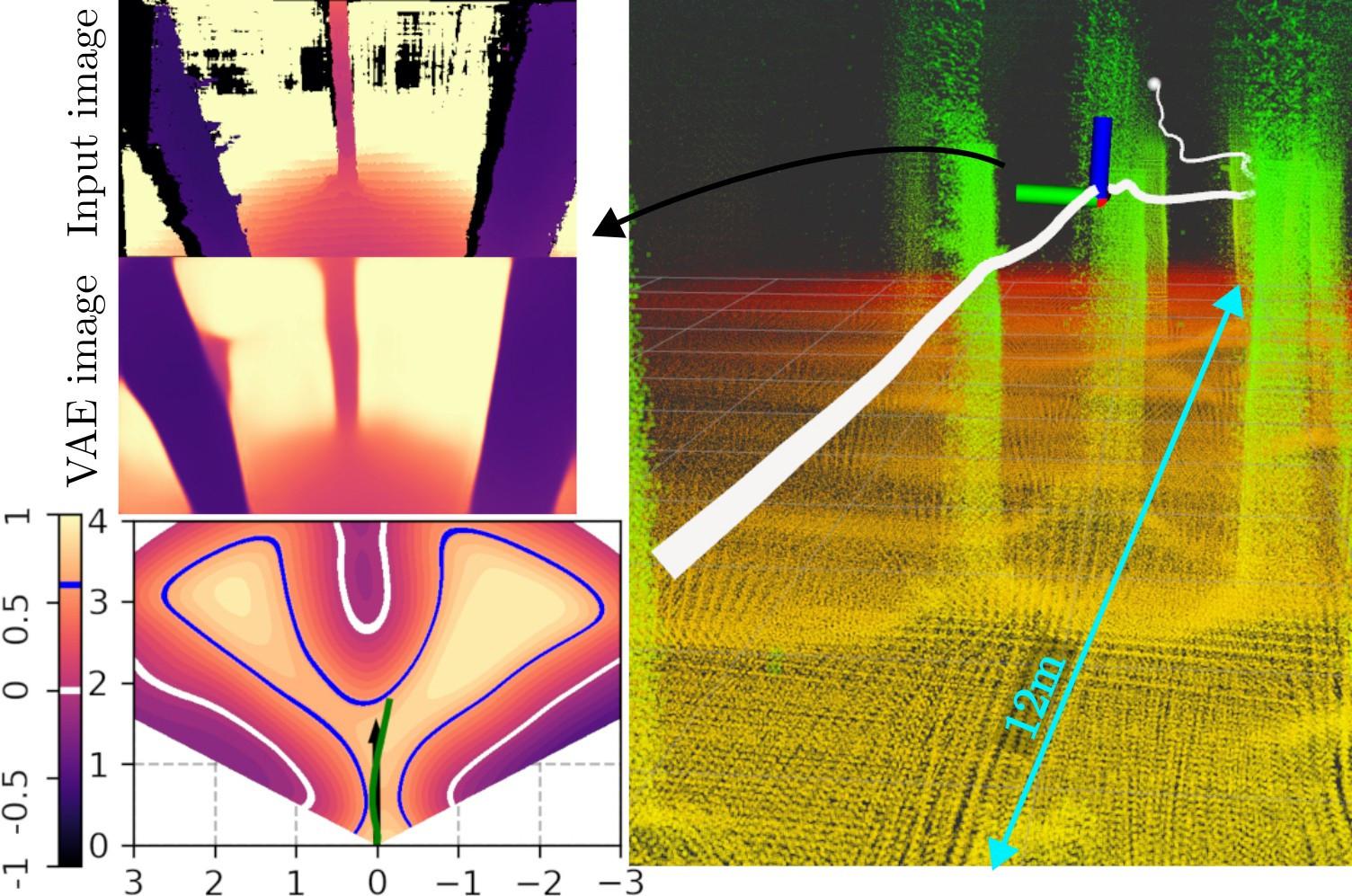}
    \caption{Third-person view of a trajectory (white path) among trees, showing the aggregated pointclouds from the depth camera.
        On the left is the input depth image at a given time instant,
        along with its VAE reconstruction,
        and a top-view of the $z_B=0$ slice of the neural SDF, its $0$-~and ($r+\epsilon$)-levelsets (respectively white and blue), the reference velocity (black arrow), and the predicted trajectory (green line).}
\label{fig:valid:xp_cam}%
\end{figure}

\begin{figure*}[t]
\centering
    \includegraphics[width=\linewidth]{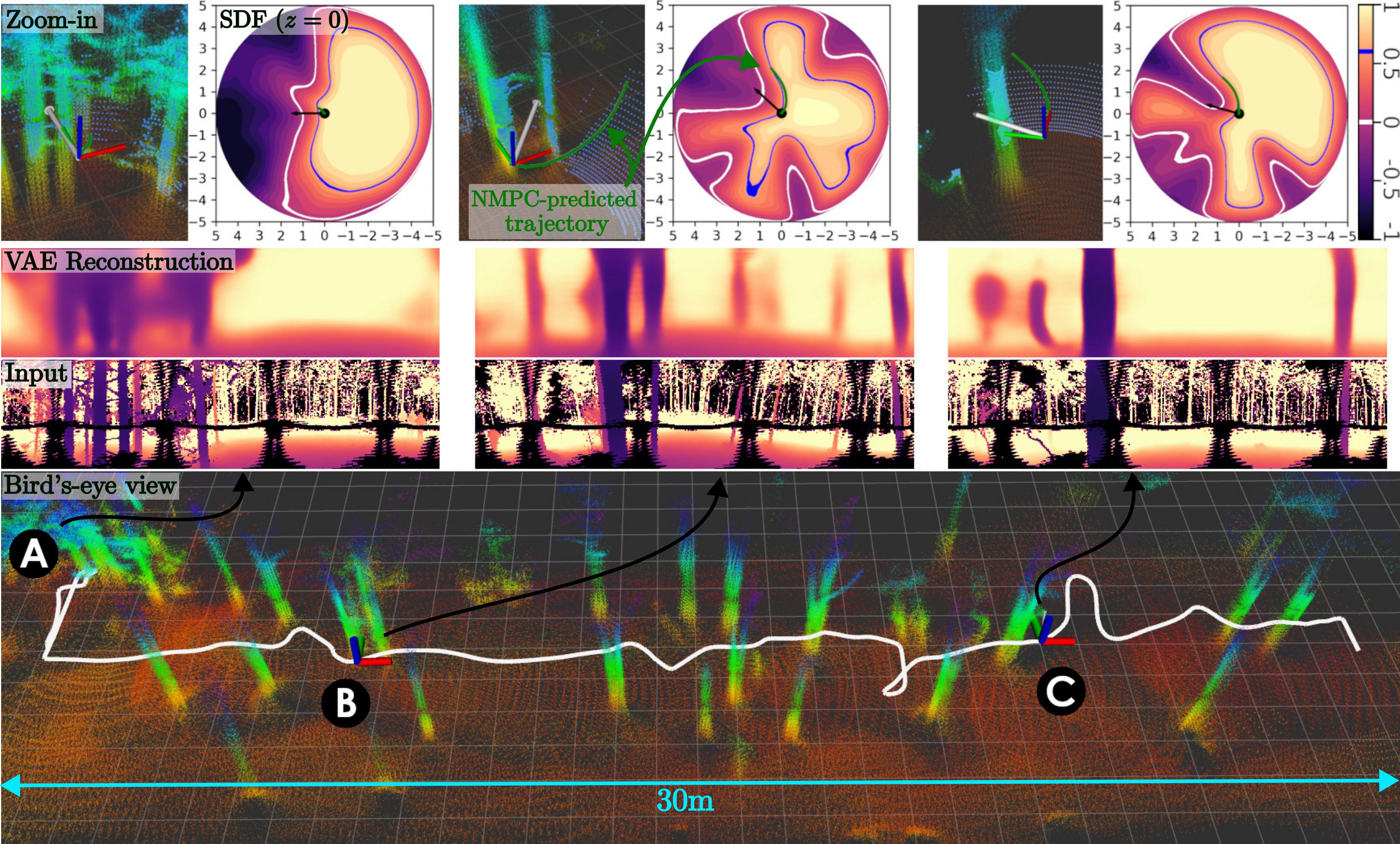}
    \caption{Bird's eye visualization of a LiDAR-based experiment among trees.
        The top-most row depicts three zoomed-in time instances A, B, and C,
        along with the $z_B=0$ slice of the neural SDF, following the same nomenclature as \fig{fig:valid:xp_cam}.
        The next two rows show the corresponding LiDAR input range image and its VAE reconstruction.}
    \label{fig:valid:xp_lidar}
\end{figure*}

We finally present hardware experiments for both the depth camera and $360$\degree LiDAR cases.
Experiments are conducted with a custom-built RMF-class quadrotor~\cite{DePetris22},
with dimensions $0.52\times 0.52\times\SI{0.3}{m}$ and mass \SI{2.58}{kg}.
The AR integrates PX4-based autopilot avionics for low-level control,
together with an NVIDIA Orin NX Single-Board Computer (SBC) running ROS Noetic.
The exteroceptive sensor used for navigation is either an Ouster OS0-64 LiDAR or a Luxonis OAK-D Pro Wide depth camera (whose halved horizontal FoV $\alpha_H$ is $63$\degree).
The sensor frequencies are set to \SI{20}{Hz}.
Further, the system employs a Texas Instruments IWR6843AOP FMCW radar sensor.
Odometry is obtained by fusing a VectorNav VN-100 IMU with the LiDAR observations using CompSLAM~\cite{Khattak20}, or with the radar measurements \cite{Nissov24} in the drifting case.
A block diagram of the system is depicted in~\fig{fig:valid:block_diag}.

The VAE encoding is executed on the Orin NX GPU, with an average inference time of \SI{12.4}{ms}.
The neural NMPC solver (using SDF${}_{256-64}$) runs on the CPU, with an average solving time of \SI{15.4}{ms}.
The resulting control frequency, including the reference velocity reference generation and the overhead induced by the Python implementation, is $\approx\SI{40}{Hz}$.

Recordings of each of the three following experiments are presented respectively in Extension 2, Extension 3, and Extension 4, along with relevant visualizations to appreciate the action of the NMPC.

\subsubsection{Experiment with Depth Camera}\label{sec:valid:xp:cam}

This first experiment is conducted using the depth camera for navigation, with LIDAR-inertial odometry.
\fig{fig:valid:xp_cam} presents an overview of the experiment.
The AR is tasked with reaching a waypoint \SI{15}{m} ahead, through a \SI{12}{m}-long tree-filled section.
The reference speed is set to $v_\text{ref}=\SI{1.5}{m/s}$,
and the observed $90$th percentile velocity is \SI{1.25}{m/s}.
The resulting motion properly avoids the trees, despite the naive environment-agnostic reference velocity.
The systematic noise in the input image (black pixels) is handled by the VAE.
The encoded SDF shows a decrease toward the obstacles, and the constraint embedded in the NMPC  (pictured as the blue line) becomes active, deflecting the trajectory to the side.

\begin{figure*}[t]
\centering
    \includegraphics[width=\linewidth]{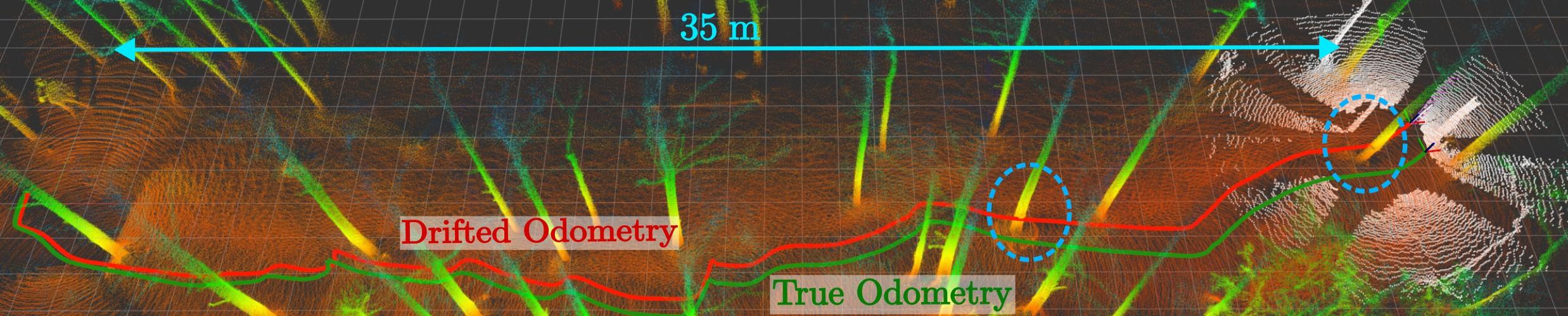}
    \caption{Bird's eye view of the trajectory with both odometries (drifted in red, ground truth in green).
        The white pointcloud depicts a single observation given to the NMPC. The controller is agnostic of the map, and operates in the local frame, and therefore maintains collision avoidance despite the drift.}
\label{fig:valid:xp_drift_odom}%
\end{figure*}

\subsubsection{Experiment with LiDAR with Adversarial Reference Velocity}\label{sec:valid:xp:lidar}

This second experiment is conducted using the LiDAR for navigation and LiDAR-inertial odometry.
Unlike the previous setup,
the reference policy is not provided by the goal-seeking planner.
Instead, an adversarial velocity input is provided by a human operator through a remote joystick, actively trying to collide the AR with the trees.
\fig{fig:valid:xp_lidar} pictures three time instances along the experiments which highlight that the NMPC effectively prevents collision by either deflecting the trajectory through some free space (instances B and C), or by bringing the robot to a full stop (instance A) when the trees form a wall-like obstacle, blocking the half-space where the velocity projection would yield a cost decrease.
Additionally, \fig{fig:valid:xp_lidar} also illustrates that the VAE is capable of successfully reconstructing the range image despite the significant amount of invalid pixels in the input image
(both in the long-range regions and within four vertical clusters obscured by the propellers and the outer cage of the AR).

\subsubsection{Experiment with Drifting Odometry}\label{sec:valid:xp:drift_odom}

This last experiment implements the drifting odometry case discussed in \sect{sec:valid:drift_odom}.
The experimental settings are similar to the previous one, \ie, the LiDAR range image is used for maintaining collision avoidance in the presence of adversarial velocity reference.
However, instead of relying on the accurate LiDAR-based odometry, we implement an Frequency Modulated Continuous Wave (FMCW) radar-based velocity estimator which provides, through integration, a drifting position and heading odometry.
Indeed, while the LiDAR-based position estimate is accurate in well-structured environments (including forests), the radar measurements have several orders of magnitude fewer points with generally greater noise, thus resulting in scan-to-scan matching approaches being impractical.
However, the availability of Doppler measurements enable reliable velocity estimation, though without position corrections, inevitably leading to drift over time.

Specifically, we implement the radar-inertial part of the factor-graph estimator presented in~\cite{Nissov24}, whose details are reported in \appen{app:radar}.
The odometry has an RPE of $3\%$ and a heading RPE of $2$\degree against the LiDAR odometry from~\cite{Khattak20}, which we use as ground truth.

\fig{fig:valid:xp_drift_odom} shows a bird's eye view of the trajectory.
Both odometries (red for radar, green for ground truth) are reported, showing the progressive drift between the two.
This experiment further verifies the associated results on drifting odometry previously presented in simulation within Section~\ref{sec:valid:drift_odom}.

\section{Conclusion}\label{sec:ccl}

This work contributes an NMPC framework for collision avoidance using a neural SDF as a differentiable, volumetric approximation of exteroceptive data.
This representation enables the optimal controller to enforce position constraints, thus enforcing collision avoidance.
The neural architecture is divided into two parts, trained sequentially:
a VAE which encodes the image in a low-dimensional latent space,
and a 4-layer coordinate-based MLP which approximates the corresponding SDF.
We further perform a theoretical analysis of the recursive feasibility and stability properties of the proposed NMPC, and derive corresponding terminal conditions.
The neural SDF encoding is first evaluated, then the NMPC controller is evaluated through several ablation studies and comparisons.
Furthermore, it is validated through outdoor experiments at \SI{1.5}{m/s}, demonstrating the avoidance of trees against adversarial inputs and drifting odometry.

This method results in a tightly integrated control and collision-avoidance policy that achieves competitive navigation performance with low computational cost.
Additionally, the approach exhibits strong resilience -- within identified thresholds -- in scenarios where odometry estimation drifts.
This addresses the primary motivation for adopting mapless navigation policies.
Finally, our results demonstrate the effectiveness of neural SDF encoding as a foundation for mapless navigation.
Future work includes seven key directions.
First, improving the framework's implementation allows for reduced sensing-to-action delay.
This could be achieved by optimizing neural network architecture and weights, and improving CPU-GPU memory transfer.
Second, investigating energy-based approaches to ensure that the NMPC retains a dissipative behavior when the active collision constraints prevent matching the reference velocity.
Third, extending the neural representation to consider the orientation would relax the spherical robot assumption and enhance maneuverability.
Fourth, incorporating visual sensing alongside or instead of range data could mitigate sensor degradation and better capture fine features, with a bimodal VAE fusing sensor inputs.
Fifth, relaxing the $0$-memory assumption through temporal observation windows, key-frame selection, and uncertainty-aware filtering.
Another avenue for this task is the use of neural world models, encoding a temporal sequence of image data into a latent representation.
Sixth, relaxing the static world assumption by predicting obstacle motion over time would further benefit from world modeling.
Finally, future work could focus on the uncertainty awareness,
w.r.t. both the noisy state information and the coarsely approximated SDF from the (also noisy) observations,
\eg by extending the proposed method to an uncertainty-informed stochastic MPC, or using a robust formulation that incorporates error bounds.

\begin{acks}
We thank Morten Nissov for his help in the setup of the radar and the corresponding estimation software.
\end{acks}

\section*{Author contributions}

\section*{Statements and declarations}
\subsection*{Ethical considerations}
This article does not contain any studies with human or animal participants.
\subsection*{Consent to participate}
Not applicable.
\subsection*{Consent for publication}
Not applicable.

\begin{dci}
The author(s) declared no potential conflicts of interest with respect to the research, authorship, and/or publication of this article.
\end{dci}

\begin{funding}
This work was partially supported by the European Commission Horizon projects DIGIFOREST (EC 101070405) and SPEAR (EC 101119774).
\end{funding}

\bibliographystyle{SageH}
\bibliography{my_bib.bib}
\appendix

\section{Training Signals}\label{app:training}

\subsection{VAE Training}

\begin{figure}[t]
\centering
        \includegraphics[width=\columnwidth]{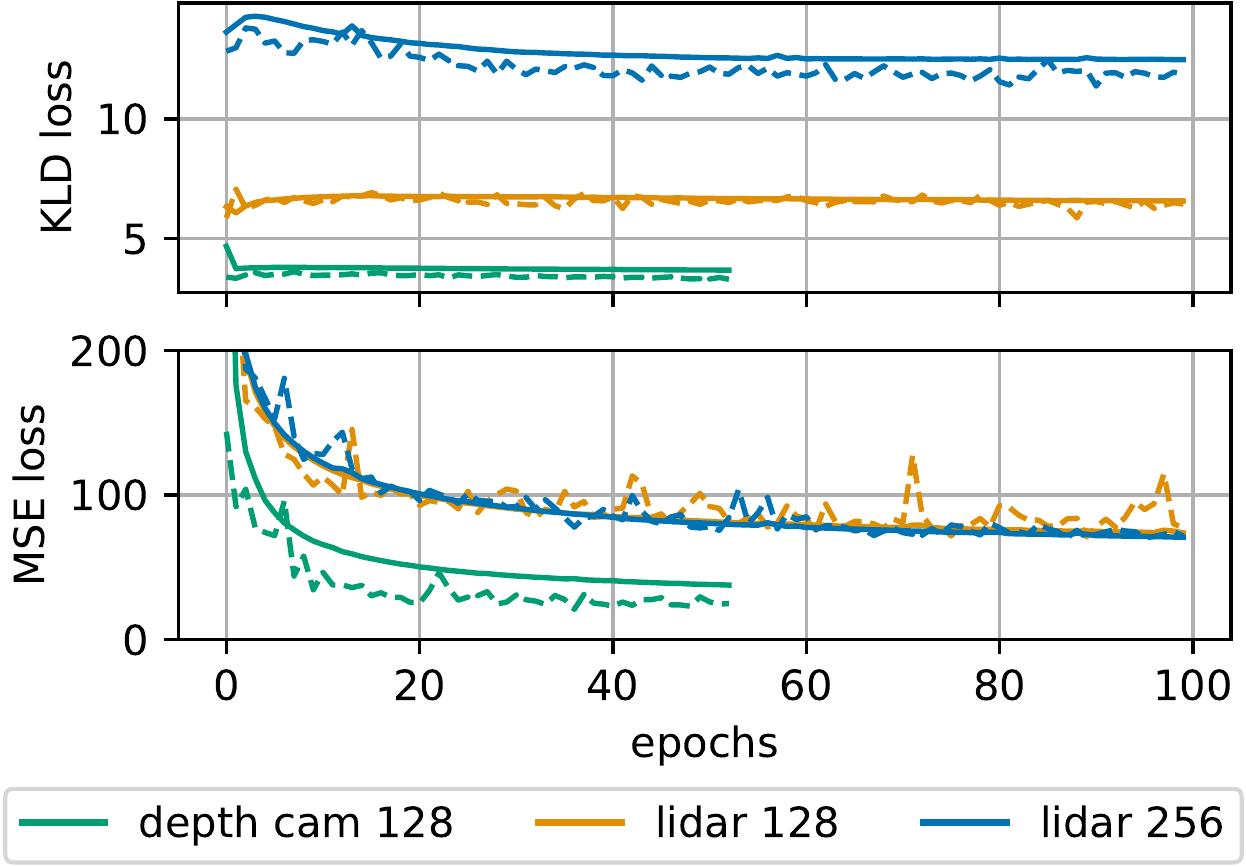}
    \caption{Training (solid lines) and validation (dashed lines) losses for 2 different VAEs input image types (LiDAR and Depth Camera).
    In the LiDAR case, we report the training signals for 2 different latent sizes.}
\label{fig:app:vae}%
\end{figure}

This section presents the training losses for the VAE networks used in the paper.
The networks are trained with a latent dimension of $128$, as recalled in \sect{sec:valid:vae}.
Figures~\ref{fig:app:vae} report the MSE and KLD losses for both depth camera and LiDAR VAEs, using the datasets presented in \tab{tab:nn:training_dataset}.

Figure~\ref{fig:app:vae} illustrates that the depth camera VAE converges to a lower MSE value than the LiDAR VAE.
This correlates with the results presented in \sect{sec:valid:vae} and is explained by the larger spatial frequency of the signal to encode.
The figure also reports the training loss for the LiDAR VAE trained with a latent dimension of $256$.
The results indicate that a latent dimension larger than $128$ brings minimal improvements to the reconstruction,
and correlates with the results reported in~\cite{Kulkarni23c}.

We note that the KLD is higher for the $256$-latent training as the $\beta$ weight is scaled as a function of the latent size~\cite{Higgins17}.

The training duration for 100 epochs is $\sim$\SI{9}{h} on an NVIDIA GeForce RTX 3090 GPU, using the full dataset from~\tab{tab:nn:training_dataset}.

\begin{figure}[t]
\centering
    \includegraphics[width=\linewidth]{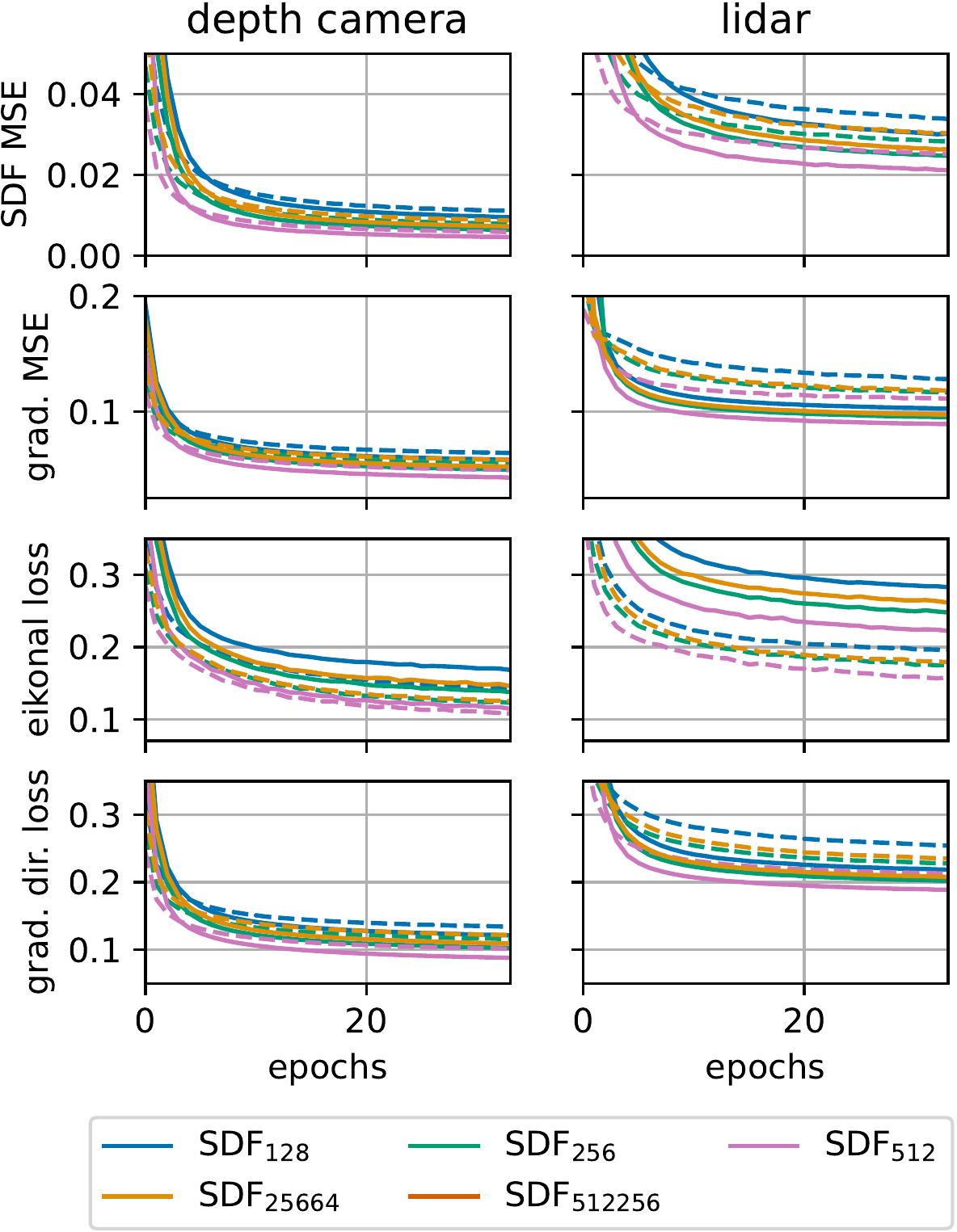}
    \caption{Color-coded validation losses and metric for the $6$ evaluated neural network sizes in Section~\ref{sec:valid:sdf}.
        The solid lines depict the signals for the training scheme proposed in Section~\ref{sec:nn:train:sdf}. Dashed lines report training signals for optimization on the eikonal loss and the gradient direction loss.}
\label{fig:app:sdf}%
\end{figure}

\subsection{SDF MLP Training}

The section presents the training losses for the various SDF networks evaluated in \sect{sec:valid:sdf}.
Figure~\ref{fig:app:sdf} reports the two training losses, \ie, the SDF MSE as well as the SDF gradient MSE.

Additionally, we report therein (as dashed lines) the training losses using the gradient direction and eikonal losses, as presented in, \eg,~\cite{Ortiz22}.

The \textit{direction+eikonal} training presents a faster convergence rate of the training signals in the early epochs compared to the \textit{MSE} training.
However, both trainings converge to similar values.

In both cases, the eikonal loss remains relatively high, in particular with smaller networks.
This is due to the complex topology of the learned SDF network, which has several angular points at which the gradient is discontinuous.
Because the neural approximation is smooth, the eikonal loss remains quite high in those regions.

The training duration for \num{40} epochs is $\sim$\SI{12}{h} on an NVIDIA GeForce RTX 3090 GPU.
We note that the main bottleneck in terms of training time is the computation of the target SDF values for the sampled points, which is largely impacted by the chosen resolution of the grid as presented in \sect{sec:nn:sdf}.
Solutions to speed up the training include pre-computing the SDF values on pre-sampled data points,
and improving the distance transform approximation algorithm by avoiding redundant computation.

\section{Choice of Dynamic Model}\label{app:dyn}

In order to evaluate the impact of different system dynamics representations in the NMPC, we conduct a comparative study using various levels of simplification for the multirotor dynamics.
Specifically, we assess:
\begin{itemize}
    \item a double integrator (2\toptext{nd} order linear system);
    \item the corresponding 2\toptext{nd} nonlinear representation with thrust/attitude mapping to the acceleration (used throughout this paper);
    \item a model extended with a simplified, 1\toptext{st} order response for the roll and pitch (as commonly done in NMPC for quadrotors, see, \eg,~\cite{Kamel17});
    \item and a 4\toptext{th} order dynamics representation, detailed, \eg, in~\cite{Lee10}.
\end{itemize}

For fair comparison, the four NMPC are tuned for good tracking of a lemniscate trajectory with high accelerations, without any perception constraints.

We design trajectories aiming to excite the higher order dynamics of the system with higher reference speed (\SI{3}{m/s}) and sharp turns in cluttered environments, similar to those presented in \sect{sec:valid:drift_odom}.
Results, summarized in \tab{tab:app:dyn}, indicate only minor improvements in success rate as model complexity increases.
This suggests that performance is more constrained by perception and the collision avoidance system than by the limits of the dynamics representation.
However, we observe that more accurate dynamics models result in smoother trajectories with fewer acceleration spikes.

Finally, as expected, we observe that the computation time increases by 9\% with the model complexity.
While this augmentation is non-negligible, it does not change the order of magnitude of the optimization complexity, confirming numerical tractability with all evaluated models.

\begin{table}[t]
\centering
    \scalebox{0.97}{
        \small
\begin{tabular}{r|cccc}
    \toprule
        &
        \textbf{\footnotesize \makecell{2\toptext{nd} order\\linear}} &
        \textbf{\footnotesize \makecell{2\toptext{nd} order\\nonlinear}} &
        \textbf{\footnotesize \makecell{2\toptext{nd} order\\nonlinear \\ with 1\toptext{st} \\ order att.}} &
        \textbf{\footnotesize \makecell{4\toptext{th} order}} \\
    \midrule
        {\footnotesize \textbf{Success}} &
        17/20 &
        17/20 &
        18/20 &
        18/20 \\
        {\footnotesize \textbf{Path excess [\%]}} &
        7.9 &
        6.2 &
        5.8 &
        4.9 \\
        {\footnotesize \textbf{Avg. velocity}} &
        2.27 &
        2.3 &
        2.28 &
        2.26 \\
        {\footnotesize \textbf{Jerk integral}} &
        21.25 &
        18.51 &
        17.03 &
        15.84 \\
        {\footnotesize \textbf{Comp. time [\unit{ms}]}} &
        10.78 &
        11.08 &
        11.29 &
        11.73 \\
    \bottomrule
\end{tabular}
\normalsize

    }
    \caption{Evaluation metrics for different dynamics models with increasing accuracy.}
\label{tab:app:dyn}%
\end{table}

This study highlights that, in the considered scenarios, the coarseness of the dynamics model is not a limiting factor.
We support this claim through two main considerations.
First, the perception constraints prevent aggressive tilting (to ensure that the trajectory remains within the FoV) and large velocities (because of the maximum range in the SDF, which enforces the predicted positions to remain within close range).
As a result, our framework effectively restricts the operational domain of the quadrotor dynamics, justifying the relevance of our approximation.
Second, as in many perception-driven tasks, the collision-avoidance system itself remains the primary performance bottleneck.
Failures to prevent collisions are more likely to result from its limitations than from the inaccuracies of the dynamics model.

Therefore, we argue that the model presented in \sect{sec:mpc:model} provides a relevant description of the dynamics for navigation without agile maneuvering.
Finally, we note that our framework can accommodate any dynamics representation -- linear or nonlinear -- to suit the specific use case.

\section{Lyapunov Stability}\label{app:stab}

This section derives the condition on the terminal cost defined in Eq.~\eqref{eq:mpc:term_cost} such that the Lyapunov stability condition~\eqref{eq:mpc:stab_condition} is satisfied.

We recall that the terminal cost is defined as

\begin{equation}
    V(\vect{x}) = p~\vect{v}\transp\vect{v},
\label{eq:stab:term_cost}%
\end{equation}
where $p > 0$,
and the Lyapunov stability is verified if

\begin{equation}
    V(\vect{x}_{N+1}) - V(\vect{x}_N)
    \le - \ell(\vect{x}_N, \pi_\text{b}(\vect{x}_N)),
\label{eq:stab:term_cost_cond}%
\end{equation}
where $\ell(\vect{x}_N, \pi_\text{b}(\vect{x}_N))$ is the stage cost, evaluated in $\vect{x}_N$ applying the braking policy~\eqref{eq:mpc:a_brake}.

Writing the discrete-time control policy for \eqref{eq:mpc:a_brake} and accounting for the edge case causing overshoot when getting close to standstill $\vect{v} = 0$,
the policy becomes:

\begin{equation}
    \pi_\text{b}(\vect{x}) =
    \begin{cases}
        \zeta(-\vect{a}_\text{b}(\vect{v}))
            \quad & \text{if} \norm{\vect{v}} > \norm{\vect{a}_\text{b}(\vect{v})}\delta t\\
        \zeta(-\frac{\vect{v}}{\delta t})
            & \text{otherwise}
    \end{cases},
\label{eq:stab:policy_brake_discrete}%
\end{equation}
where $\zeta$ denotes the nonlinear mapping from the acceleration to the thrust and attitude input~\eqref{eq:mpc:dyn:zeta},
and $\delta t$ is the discretization time step.

The left-hand side in Eq.~\eqref{eq:stab:term_cost_cond} is therefore written

\begin{equation}
    \begin{aligned}
        V(\vect{x}_{N+1}) - V(\vect{x}_N) &= p (\norm{\vect{v}_{N+1}}^2 - \norm{\vect{v}_N}^2) \\
        ~ &= - p \Delta v (2\norm{\vect{v}_N} - \Delta v)
    \end{aligned}\quad,
\label{eq:stab:term_cost_vN}%
\end{equation}
where $\Delta v = \norm{\vect{v}_N} - \norm{\vect{v}_{N+1}}$ is the absolute velocity decrease induced by applying $\pi_\text{b}(\vect{x}_N)$.
By construction of this maximum braking policy, a constant deceleration is applied such that the velocity decrease can be expressed by forward Euler integration over $\delta t$.
Thus, by Eq.~\eqref{eq:stab:policy_brake_discrete} it holds

\begin{equation}
    \Delta v = 
    \begin{cases}
        \norm{\vect{a}_\text{b}(\vect{v}_N)}\delta t
            \quad & \text{if} \norm{\vect{v}_N} > \norm{\vect{a}_\text{b}(\vect{v}_N)}\delta t \\
        \norm{\vect{v}_N}
            & \text{otherwise}
    \end{cases}.
\label{eq:stab:dv}%
\end{equation}

Hence, Eq.~\eqref{eq:stab:term_cost_vN} becomes

\scriptsize
\begin{equation}
\begin{aligned}
    &V(\vect{x}_{N+1}) - V(\vect{x}_N) = \\
    &\begin{cases}
        - p \norm{\vect{a}_\text{b}(\vect{v}_N)}\delta t (2\norm{\vect{v}_N} - \norm{\vect{a}_\text{b}(\vect{v}_N)}\delta t)
            \quad & \text{if} \norm{\vect{v}_N} > \norm{\vect{a}_\text{b}(\vect{v}_N)}\delta t \\
        - p\norm{\vect{v}_N}^2
            & \text{if} \norm{\vect{v}_N} \le \norm{\vect{a}_\text{b}(\vect{v}_N)}\delta t
    \end{cases}
\label{eq:stab:cond_cases}%
\end{aligned}
\end{equation}
\normalsize
We consider each case separately.

\subsubsection*{First case: $\norm{\vect{v}_N} > \norm{\vect{a}_\text{b}(\vect{v}_N)}\delta t$}

Considering the first case, inserting the inequality $\norm{\vect{v}_N} > \norm{\vect{a}_\text{b}(\vect{v}_N)}\delta t$ in Condition~\eqref{eq:stab:term_cost_cond} and reformulating yields

\begin{equation}
    p \norm{\vect{a}_\text{b}(\vect{v}_N)}^2\delta t^2 \ge \ell(\vect{x}_N, \pi_\text{b}(\vect{x}_N)).
\label{eq:stab:case1:cond}%
\end{equation}

The left-hand side of the inequality is upper-bounded since $\norm{\vect{a}_\text{b}(\vect{v})}$ is bounded due to input constraints.
Further, $\ell$ is quadratic in $\norm{\vect{v_N}}$, Eq.~\eqref{eq:stab:case1:cond} can not be verified globally with any finite $p$.
Instead, we find $p$ such that the inequality holds for all $\vect{x}_N$ within a predefined area of operation since arbitrarily large velocities are not intended states for autonomous navigation. Specifically, we want the inequality \eqref{eq:stab:case1:cond} to hold for $\norm{\vect{v}_N} \le \overline{v}$.
Assuming that the reference velocity is also bounded,
an upper bound on the stage cost $\overline{\ell}$ can be expressed.

Similarly, a lower bound on $\norm{\vect{a}_\text{b}(\vect{v})}$, denoted $\underline{\vect{a}_\text{b}} > 0$, can be obtained as

\begin{equation}
\begin{aligned}
    \underline{\vect{a}_\text{b}} &= \min_{\vect{v}} \norm{\vect{a}_\text{b}(\vect{v})} \\
    &s.t. \quad \norm{\vect{v}} \le \overline{v}
\end{aligned}
\end{equation}

Rewriting Eq.~\eqref{eq:stab:case1:cond}, the corresponding condition on $p$ is thus given by

\begin{equation}
    p \ge \frac
        {\overline{\ell}}
        {\underline{\vect{a}_\text{b}}^2\delta t^2}.
\label{eq:stab:case1:p}%
\end{equation}

\subsubsection*{Second case: $\norm{\vect{v}_N} \le \norm{\vect{a}_\text{b}(\vect{v})}\delta t$}

The second case in Eq.~\eqref{eq:stab:cond_cases} corresponds to the specific case when the velocity approaches the discontinuity in $\pi_\text{b}$ for $\vect{v} = 0$.

In this case, Condition~\eqref{eq:stab:term_cost_cond} becomes

\begin{equation}
     p \norm{\vect{v}_N}^2 \geq
        \norm{\mat{Q}}_2 \norm{\vect{v}_N - \vect{v}_\text{ref}}^2 +
        \norm{\pi_\text{b}(\vect{x}_N)}^2_{\mat{R}}.
        \label{eq:stab:case2:bound_2}
\end{equation}

For a nonzero $\vect{v}_\text{ref}$, the above inequality cannot hold since the left-hand side vanishes for $\vect{v}_N$ approaching zero, while the term $\norm{\vect{v}_N - \vect{v}_\text{ref}}^2$ does not.
In this case, the competing objectives in the terminal cost (reduce velocity to zero) and stage cost (track $\vect{v}_\text{ref}$) do not allow for showing asymptotic stability without further assumptions.
In this case, we assume the presence of a reference governor, which modifies $\vect{v}_\text{ref}$ in a suitable manner to retain stability.
This can easily be achieved by setting $\vect{v}_\text{ref} = 0$ for all $t_i \geq t_N$ in the case when $\norm{\vect{v}_N} \le \norm{\vect{a}_\text{b}(\vect{v})}\delta t$.
This simple modification of future $\vect{v}_\text{ref}$ essentially results in a reduction of the reference velocity beyond the prediction horizon, whenever $\norm{\vect{v}_N} \le \norm{\vect{a}_\text{b}(\vect{v})}\delta t$.

Inserting $\vect{v}_\text{ref} = 0$ into \eqref{eq:stab:case2:bound_2} results in

\begin{equation}
     p \norm{\vect{v}_N}^2 \geq
        \norm{\mat{Q}}_2 \norm{\vect{v}_N}^2 +
        \norm{\pi_\text{b}(\vect{x}_N)}^2_{\mat{R}}.
        \label{eq:stab:case2:bound_3}
\end{equation}

Inserting the Cauchy-Schwarz inequality, an upper bound of the stage cost can be found as

\begin{equation}
    \ell(\vect{x}_N, \pi_\text{b}(\vect{x}_N)) \le
        \norm{\mat{Q}} \norm{\vect{v}_N}^2 +
        \norm{\pi_\text{b}(\vect{x}_N)}^2_{\mat{R}},
\label{eq:stab:case2:bound_l}%
\end{equation}
where $ \norm{\mat{Q}}$ is the 2-norm of the diagonal matrix $\mat{Q}$.

From Eq.~\eqref{eq:stab:policy_brake_discrete}, we have that $\pi_\text{b}(\vect{x}_N) = \zeta(\frac{\vect{v}_N}{\delta t})$.
Since the mapping $\zeta$ is nonlinear in $\vect{v}_N$, the term $\norm{\pi_\text{b}(\vect{x}_N)}^2_{\mat{R}}$ is not simply quadratic in $\vect{v}_N$.
Instead, we find a quadratic upper bound on $\norm{\pi_\text{b}(\vect{x}_N)}^2_{\mat{R}}$ by defining a scalar $\tilde{r} > 0$ such that

\begin{equation}
    \norm{\zeta(\frac{\vect{v}_N}{\delta t})}^2_{\mat{R}} \le
        \tilde{r} \norm{\vect{v}_N}^2.
\label{eq:stab:case2:R_tilde}%
\end{equation}

Such scalar $\tilde{r}$ is obtained by inverting $\zeta$ and finding the maximum over braking actions $\bmat{T,~\phi,~\theta}$,
\ie

\begin{equation}
    \tilde{r} = \max_{T,~\phi,~\theta} \frac
        {m^{2} \left(r_{1} \phi^{2} + r_{2} \theta^{2} + r_{3} \left(T - g m\right)^{2}\right)}
        {\delta_{t}^{2} \left(T^{2} - 2 T g m \cos{\phi} \cos{\theta} + g^{2} m^{2}\right)},
\end{equation}
where $\mat{R} = \text{diag}(r_1, r_2, r_3)$.

Thus, a sufficient condition for Condition~\eqref{eq:stab:term_cost_cond} to hold is

\begin{equation}
    p \ge \norm{\mat{Q}} + \tilde{r}.
\label{eq:stab:case2:p}%
\end{equation}

Combining the first and second cases, a sufficient condition to ensure the Lyapunov stability condition~\eqref{eq:mpc:stab_condition} is provided by selecting the terminal cost

\begin{equation}
    V(\vect{x}_N) = \max
        \left(
            \frac
                {\overline{\ell}}
                {\underline{\vect{a}_\text{b}}^2\delta t^2},~
            \norm{\mat{Q}} + \tilde{r}
        \right) ~\vect{v}_N\transp\vect{v}_N.
\label{eq:stab:p}%
\end{equation}

\section{Radar-Inertial Odometry}\label{app:radar}

The radar-inertial odometry follows~\cite{Nissov24}, but with slight enhancements.
The core structure is as follows: a factor graph-based smoother that fuses Frequency Modulated Continuous Wave (FMCW) radar pointclouds and IMU measurements of angular velocity and specific force.
For each radar measurement, a node is created in the graph and connected to the graph with an IMU pre-integration factor, following~\cite{Forster17}.
Afterwards, a radial speed factor, linking the navigation states to the radar Doppler measurements, is connected to the newly created node.
To maintain computational efficiency, the graph size is limited by duration, and sufficiently old nodes are periodically marginalized following the approach in~\cite{GTSAM}.

Furthermore, a secondary integration routine provides IMU-rate odometry estimates for use by the proposed method.
This architecture enables high-rate, accurate velocity estimation, with inevitable position drift resulting from the integration of noisy signals.

\section{Index to multimedia Extensions}\label{app:multimedia}

\begin{table}[h]
\centering
    \small
\newlength{\descwidth}
\setlength{\descwidth}{5cm}
\begin{tabular}{ccp{\descwidth}}
    \toprule
        \textbf{Extension} & \textbf{\makecell{Media \\ type}} & \textbf{Description} \\
    \midrule
        1 & Video & A visualization of both the ablation and comparative studies presented in Sections \ref{sec:valid:ablation} and \ref{sec:valid:comparison}, providing insights into the evaluated environments. \\
        2 & Video & The recording and relevant visualizations of the experiment presented in \sect{sec:valid:xp:cam}, where the constrained-FoV depth camera is used for collision avoidance. \\
        3 & Video & The recording and relevant visualizations of the experiment presented in \sect{sec:valid:xp:lidar}, where a human operator provides adversarial velocity reference to the controller. \\
        4 & Video & The presentation of the simulations conducted in Section~\ref{sec:valid:drift_odom}, as well as the recording and relevant visualizations of the experiment presented in Section~\ref{sec:valid:xp:drift_odom}, where a radar-based velocity estimator is used to assess resilience to position drift \\
    \bottomrule
\end{tabular}
\normalsize

    \caption{Index of multimedia extensions.}
\label{tab:extensions}
\end{table}

\end{document}